\documentclass[10pt,journal,compsoc]{IEEEtran}
\usepackage{graphicx}
\usepackage{amsmath,amssymb} 

\usepackage[pagebackref=true,breaklinks=true,letterpaper=true,colorlinks,bookmarks=false]{hyperref}

\usepackage{amsmath}
\usepackage{times}
\usepackage{helvet}
\usepackage{courier}
\usepackage{graphicx}
\usepackage{bm}
\usepackage{amsmath,amssymb} 
\usepackage{color}
\usepackage{bbm}
\usepackage{epstopdf}
\usepackage{caption}
\usepackage{subcaption}
\usepackage{enumitem}
\usepackage{calc}
\usepackage{multirow}
\usepackage{xspace}
\usepackage{booktabs}
\usepackage{mathrsfs}
\usepackage{mathtools}
\usepackage{mathptmx}
\usepackage{array}
\usepackage[font=small,skip=4pt]{caption}
\usepackage{graphicx}
\usepackage[switch]{lineno}
\usepackage{threeparttable}

\newcommand{\figref}[1]{Fig\onedot~\ref{#1}}
\newcommand{\equref}[1]{Eq\onedot~\eqref{#1}}
\newcommand{\secref}[1]{Sec\onedot~\ref{#1}}
\newcommand{\tabref}[1]{Tab\onedot~\ref{#1}}

\newcommand{\ve}[1]{{\mathbf #1}} 

\newcommand{\peng}[1]{\textcolor{black}{#1}}

\newcommand{\by}[2]{\ensuremath{#1 \! \times \! #2}}
\newcommand{\thickhline}{%
    \noalign {\ifnum 0=`}\fi \hrule height 1pt
    \futurelet \reserved@a \@xhline
}

\DeclareRobustCommand\onedot{\futurelet\@let@token\@onedot}
\def\onedot{\ifx\@let@token.\else.\null\fi\xspace}
\def\eg{\emph{e.g., }}

\def\ie{\emph{i.e., }}

\def\etc{\emph{etc}\onedot} 

\def\wrt{w.r.t\onedot} 

\def\etal{\emph{et al.}}

\makeatletter
\newcommand\footnoteref[1]{\protected@xdef\@thefnmark{\ref{#1}}\@footnotemark}
\makeatother

\ifCLASSOPTIONcompsoc
  \usepackage[nocompress]{cite}
\else
  \usepackage{cite}
\fi
\ifCLASSINFOpdf
\else
\fi
\graphicspath{./fig/}

\begin{document}
%
\title{Learning Depth with Convolutional Spatial Propagation Network}

%
%

\author{~Xinjing Cheng,
        ~Peng~Wang
        and ~Ruigang Yang,~\IEEEmembership{Senior Member,~IEEE}
\IEEEcompsocitemizethanks{
\IEEEcompsocthanksitem X. Cheng, P. Wang,  R. Yang are with Baidu Research, Baidu Inc., Beijing, China. }
\IEEEcompsocitemizethanks{\IEEEcompsocthanksitem Corresponding author: P. Wang $<$wangpeng54@baidu.com$>$} }

\IEEEtitleabstractindextext{%
\begin{abstract}
In this paper, we propose the convolutional spatial propagation network (CSPN) and demonstrate its effectiveness for various depth estimation tasks. 
CSPN is a simple and efficient linear propagation model, where the propagation is performed with a manner of recurrent convolutional operations, in which the affinity among neighboring pixels is learned through a deep convolutional neural network (CNN). Compare to the previous state-of-the-art (SOTA) linear propagation model, \ie~spatial propagation networks (SPN), CSPN is $\bf2$ to $\bf5\times$ faster in practice.
We concatenate CSPN and its variants to SOTA depth estimation networks, which significantly improve the depth accuracy. Specifically, we apply CSPN to two depth estimation problems: depth completion and stereo matching, in which we design modules which adapts the original 2D CSPN to embed sparse depth samples during the propagation, operate with 3D convolution and be synergistic with spatial pyramid pooling.
In our experiments, we show that all these modules contribute to the final performance. For the task of depth completion, our method reduce the depth error over $\bf30\%$ in the NYU v2 and KITTI datasets. For the task of stereo matching, our method currently ranks $\bf1st$ on both the KITTI Stereo 2012 and 2015 benchmarks. 
The code of CSPN will be released at \url{https://github.com/XinJCheng/CSPN}.


\end{abstract}
}

\maketitle

\IEEEraisesectionheading{\section{Introduction}\label{sec:intro}}
\IEEEPARstart{D}epth estimation from images, \ie predicting absolute per-pixel distance to the camera has many applications in practice, such as augmented reality (AR), autonomous driving~\cite{chen2015deepdriving},  robotics~\cite{murray2000using,biswas2011depth,haque2017obstacle}. 
It also serves as a foundation to support other computer vision problems, such as 3D reconstruction~\cite{bascle1993stereo,zhang2015meshstereo} and recognition~\cite{xu2013epipolar,qi20173d}. 

\begin{figure*}[t]
\centering
\includegraphics[width=1.00\linewidth]{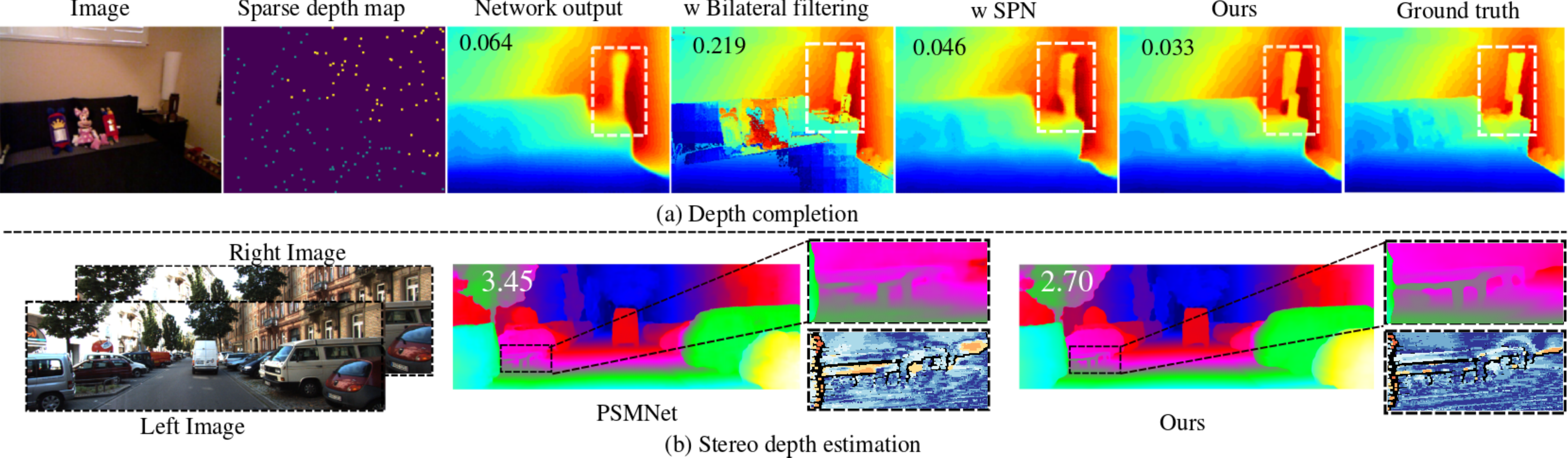}
\caption{Two depth estimation tasks focused in this paper: (a) depth completion, the output from various methods and color means the redder the further. ``Network output'' is from Ma \etal ~\cite{Ma2017SparseToDense}; ``w'' is short for with. ``w SPN'' means after refinement with SPN~\cite{liu2017learning}. The root mean square error (RMSE) is put at the left-top of each predicted depth map. 
(b) stereo depth estimation, the output from PSMNet~\cite{chang2018pyramid} and our prediction. Color means the bluer the further. The D1 error rate is put at left-top of the predictions. A significantly improved region is highlighted with dash box, and corresponding error map are shown below (the bluer the lower of error). In both cases, our outputs are significantly better. }
\label{fig:example}
\end{figure*}

As illustrated in \figref{fig:example}, in this paper, we are particularly interested in two depth estimation tasks that are practical in real applications: depth completion and stereo matching. 

\noindent\textbf{Depth completion} (\figref{fig:example}(a)), \textit{a.k.a.} sparse to dense depth conversion~\cite{LiaoHWKYL16} is a task of converting sparse depth samples to a dense depth map given the corresponding image~\cite{LiaoHWKYL16,Ma2017SparseToDense}.  It can be widely applied in robotics and autonomous driving, where depth perception is often acquired through LiDAR~\cite{velodyne}, that typically generates sparse but accurate depth measurement. By combining the sparse measurements with a reference image, we could generate a full-frame dense depth map.

For this task, recent efforts have yielded high-quality outputs by taking advantage of deep fully convolutional neural networks~\cite{eigen2015predicting,laina2016deeper} and a large amount of training data from indoor~\cite{silberman2012indoor,xiao2013sun3d,Matterport3D} and outdoor environments~\cite{geiger2012we,wang2016torontocity,huang2018apolloscape}. 
The improvement lies mostly in more accurate estimation of global scene layout and scales with advanced networks, such as VGG~\cite{simonyan2014very} or ResNet~\cite{HeZRS15}, and better local structure recovery through deconvolution operations~\cite{long2015fully}, skip-connections~\cite{ronneberger2015u} or up-projection~\cite{laina2016deeper}. 
Nevertheless, upon a closer inspection of the outputs from a contemporary approach for depth completion~\cite{Ma2017SparseToDense} (\figref{fig:example}(a) 3rd column), the predicted depth map is still blurry and often do not align well with the given image structure such as object silhouette. This motivates us to adopt linear propagation models~\cite{barron2016fast,liu2017learning} for tackling the issue.

However, directly refining with bilateral filtering using pixel RGB (\figref{fig:example}(a) 4th column) omits global image context, which produces noisy depth estimation. Most recently, Liu \etal~\cite{liu2017learning} propose to learn the image-dependent affinity through a deep CNN with spatial propagation networks (SPN), yielding better results comparing to the manually designed affinity on image segmentation. 
However, its propagation is performed in a scan-line or scan-column fashion, which is serial in nature. For instance, when propagating left-to-right, pixels at right-most column must wait the information from the left-most column to update its value.

Intuitively, depth refinement should not be order dependent. As such, we propose Convolutional Spatial Propagation Networks (CSPN), where the depth values at all pixels are updated simultaneously within a local convolutional context. The long range context is obtained through a recurrent operation. \figref{fig:example}(a) shows an example, the depth estimated from CSPN (6th column) is more accurate than that from SPN (5th column).
In our experiments, our parallel update scheme leads to significant performance improvement in both speed and quality over the serial ones such as SPN.

More specifically, in this paper, we consider three important requirements for a depth completion algorithm: 
(1) The dense depth map recovered should align with image structures;
(2) The depth value from the sparse samples should be preserved, since they are usually from a reliable sensor; and
(3) The transition between sparse depth samples and their neighboring depths should be smooth and unnoticeable.
In order to satisfy those requirements, we first add mirror connections based on the network from~\cite{Ma2017SparseToDense} to learn better image structural features. Then, we embed the sparse depth samples into the propagation process in order to keep the depth value at sparse points. Finally, our CSPN recovers image structures in detail and satisfies the transition smoothness requirements. 
In our experiments, we show our recovered depth map with just 500 depth samples produces much more accurately estimated scene layouts and scales than other SOTA strategies. 


\noindent\textbf{Stereo Matching} (\figref{fig:example}(b)), where a pair of calibrated images, commonly from a stereo camera~\cite{realsense}, are used as inputs. Specifically, for pixel $(x, y)$ in the reference image (1st column), if its corresponding disparity is $d_{x,y}$, then the depth of this pixel could be calculated
by $\frac{f * B}{d_{(x,y)}}$, where $f$ is the camera's focal length and $B$ is the distance between two camera centers. 

Current SOTA methods for stereo also rely on the advancement of deep networks~\cite{zbontar2016stereo,kendall2017end,chang2018pyramid}. Most recently, GCNet~\cite{kendall2017end} learns to incorporate geometrical context directly from the data, employing 3D Convolutions (3DConv) over the disparity cost volume.
PSMNet~\cite{chang2018pyramid} adopts a similar idea, while induces extensions at \textit{scale space}  using spatial feature pooling~\cite{he2014spatial} at the end of feature encoder and multi-scale outputs from their stacked hourglass networks~\cite{newell2016stacked} with 3DConv. 
This motivates us to lift the 2D spatially propagating CSPN to 3D, where information can also propagate within the \textit{disparity space} and \textit{scale space}, yielding more accurate estimated results because of better details and less noises (as shown in \figref{fig:example}(b) 4th column). 

Last but not least, we regard the commonly used spatial pyramid pooling (SPP)~\cite{he2014spatial} and atrous SPP (ASPP)~\cite{ChenPSA17} for multi-scale feature integration as special cases of CSPN.  With such a perspective, we propose to learn an affinity matrix for feature aggregation within each pyramid and across pyramids, yielding further performance boost in both tasks than their original variants.


We perform various experiments to validate our approaches over several popular benchmarks for depth estimation. 
For depth completion, NYU v2~\cite{silberman2012indoor} and KITTI~\cite{geiger2012we} are adopted. In both datasets, our approach is significantly better (about $30\%$ relative improvements in most key measurements) than previous deep learning based state-of-the-art algorithms~\cite{LiaoHWKYL16,Ma2017SparseToDense}. 
More importantly, it runs efficiently, yielding up to 5$\times$ acceleration over SPN on large images.
For stereo depth estimation, the Scene Flow~\cite{mayer2016large} and KITTI Stereo datasets~\cite{geiger2012we,Menze2018JPRS} are adopted, and we rank the $1_{st}$ on both the KITTI Stereo 2012 and 2015 benchmarks~\footnote{\url{http://www.cvlibs.net/datasets/kitti/eval_stereo_flow.php?benchmark=stereo}}. 

In summary, this paper has the following contributions: 
\begin{enumerate}
    \item We propose convolutional spatial propagation networks (CSPN) that learn the affinity directly from images. It has shown to be more efficient and accurate for depth estimation than the previous SOTA propagation strategy~\cite{liu2017learning}, without sacrificing the stability of linear propagation.
    
    \item  We extend CSPN to the task of depth completion by embedding the provided sparse depth samples into the propagation process, which also guarantees that the sparse input depth values are preserved in the final depth map.
    
    \item We lift CPSN to 3D CSPN for stereo depth estimation, which explores the correlation within both discrete disparity space and scale space. It helps the recovered stereo depth generate more details and avoid error matching from noisy appearance caused by sunlight or shadows \etc. 
    
    \item We provide a CSPN perspective for spatial pyramid pooling (SPP), and propose a more effective SPP module, which further boosts performance for both depth completion and stereo.
    
\end{enumerate}

The structure of this paper is organized as follows. We provide related work in~\secref{sec:related}. \secref{sec:CSPN_Module} elaborates the design, theoretical background of CSPN, and its variants including 3DCSPN and its relation with SPP.
In~\secref{subsec:sparse2dense} and \secref{subsec:stereo}, we present the details about how we adapt CSPN to depth completion and stereo matching correspondingly. Finally, we evaluate the results of our algorithms on all the tasks quantitatively and qualitatively in Sec.~\ref{sec:exp}.


\section{Related Work}
\label{sec:related}
Depth estimation has been a center problem for computer vision and robotics for a long time. Due to space limitation, here we summarize those works in several most relevant aspects.


\noindent\textbf{Single view depth estimation via CNN and CRF.} We first review single view depth estimation since methods in this area motivated our design. Deep neural networks (DCN) provide strong feature representation for various vision tasks. Depending on DCN, numerous algorithms are developed, including supervised methods~\cite{wang2015designing,eigen2015predicting,laina2016deeper,li2017two}\cite{luo2018single},\cite{jiao2018look},\cite{Fu_2018_CVPR}
semi-supervised methods~\cite{kuznietsov2017semi} or unsupervised methods~\cite{godard2016unsupervised,zhou2017unsupervised,yang2018aaai,yang2018lego,luo2018every,Jiang_2018_ECCV}.
Others tried to improve the estimated details further by appending a conditional random field (CRF)~\cite{DBLP:conf/cvpr/WangSLCPY15,Liu_2015_CVPR,li2015depth,xu2017multi,xu2018structured,Heo_2018_ECCV} and multi-task correlation with joint training~\cite{peng2016depth,jiao2018look,Qi_2018_CVPR,Xu_2018_CVPR}. 
However, the affinity for measuring the coherence of neighboring pixels is commonly manual-designed based on color similarity or intervening contour~\cite{shi2000normalized} with RBF kernel~\cite{peng2016depth,Heo_2018_ECCV}. 
In our case, CSPN learn data dependent affinity that has demonstrated to be more robust in practice.

\noindent\textbf{Depth Enhancement.}
Traditionally, depth output can be also efficiently enhanced with explicitly designed affinity through image filtering~\cite{barron2016fast,matsuo2015depth}, or data-driven ones through total variation (TV)~\cite{ferstl2013image,ferstl2015variational} and learning to diffuse~\cite{liu2016learning} by incorporating more priors into diffusion partial differential equations (PDEs).
However, due to the lack of an effective learning strategy, they are limited for large-scale complex visual enhancement.

Recently, deep learning based enhancement yields impressive results on super resolution of both images~\cite{dong2014learning,yang2014color} and depths~\cite{song2016deep,hui2016depth,kwon2015data,riegler2016atgv}. The network takes low-resolution inputs and output the high-resolution results, and is trained end-to-end where the mapping between input and output is implicitly learned.
However, these methods are trained and experimented only with perfect correspondent ground-truth low-resolution and high-resolution depth maps and often a black-box model. In our scenario, neither the input nor ground truth depth are non-perfect, \eg depths from a low cost LiDAR or a network, thus an explicit diffusion process to guide the enhancement such as SPN is necessary.

\noindent\textbf{Learning affinity for spatial diffusion.}
Learning affinity matrix with deep CNN for diffusion or spatial propagation receives high interests in recent years due to its theoretical supports and guarantees~\cite{weickert1998anisotropic}.
Maire \etal~\cite{maire2016affinity} trained a deep CNN to directly predict the entities of an affinity matrix, which demonstrated good performance on image segmentation. However, the affinity is followed by an independent non-differentiable solver of spectral embedding, it can not be supervised end-to-end for the prediction task. Bertasius \etal~\cite{bertasius2016convolutional} introduced a random walk network that optimizes the objectives of pixel-wise affinity for semantic segmentation. Nevertheless, their affinity matrix needs additional supervision from ground-truth sparse pixel pairs, which limits the potential connections between pixels. Chen \etal~\cite{chen2016semantic} try to explicit model an edge map for domain transform to improve the performance. 

The most related work is SPN~\cite{liu2017learning}, where the learning of a large affinity matrix for diffusion is converted to learning a local linear spatial propagation, yielding a simple yet effective approach for output enhancement. However, as mentioned in~\secref{sec:intro}, depth enhancement commonly needs local context, especially when sparse depth samples are available for depth completion. It might not be necessary to update a pixel by scanning the whole image. In our experiments, the proposed CSPN is more efficient and provides much better results.


\begin{figure*}[t]
\includegraphics[width=\textwidth]{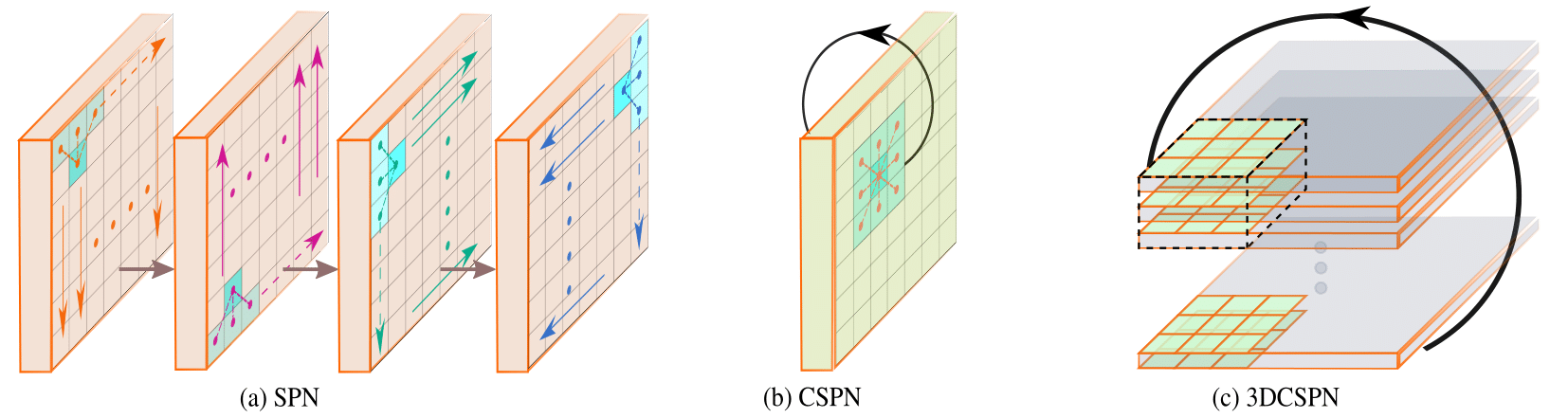}
\caption{Comparison between the propagation process in (a) SPN~\cite{liu2017learning}, (b) 2D CPSN and (c) 3D CSPN in this work. Notice for 3D CSPN, the dashed volume means one slice of the feature channel in a 4D volume with size of $d \times h \times w \times c$ (Detailed in \secref{subsec:stereo}).}
\label{fig:compare}
\vspace{-1.1\baselineskip}
\end{figure*}

\noindent\textbf{Depth completion with sparse samples.}
 Different from depth enhancement, the provided depths are usually from low-cost LiDAR, yielding a map with valid depth in only few hundreds of pixels, as illustrated in \figref{fig:example}(a).
Most recently, Ma \etal~\cite{Ma2017SparseToDense,ma2018self} propose to treat the sparse depth map as an additional input to a ResNet~\cite{laina2016deeper} based depth predictor, producing superior results than the depth output from CNN with sole image input. However, the output results are still blurry, and do not satisfy our requirements of depth as discussed in \secref{sec:intro}. In CSPN for depth completion, we embed the sampled depth samples in the diffusion process, where all the requirements we proposed are held properly.

Some other works directly convert 3D points from LiDAR to dense ones without image input~\cite{Zimmermann2017Learning,Ladicky_2017_ICCV,uhrig2017sparsity}, In these methods, the density of valid depth must be high enough to reveal the scene structure.
Other works~\cite{Zhang_2018_CVPR} complete depth with an RGB image in a manner of in-painting, where dense depth values are partially available.  These approaches are dealing with a different setup from ours.

\noindent\textbf{Stereo depth estimation.}
Stereo depth estimation has long been a central problem in computer vision. Traditionally, Scharstein and Szeliski~\cite{scharstein2002taxonomy} provide a taxonomy of stereo algorithms including: matching cost calculations, matching cost aggregation, disparity calculation and disparity refinement\cite{hirschmuller2008stereo,heise2013pm,wang2008stereoscopic}. 

CNNs were first introduced to stereo matching by Zbontar and LeCun~\cite{zbontar2016stereo} to replace the computation of the matching cost. Their method showed that by using CNNs, the matching could be more robust, and achieved SOTA results over KITTI Stereo benchmarks. However, the networks are still shallow, and it needs post-processing for refinement. 
Following~\cite{zbontar2016stereo}, several methods were proposed to increase computational efficiency~\cite{feng2017efficient,luo2016efficient}, or matching cost accuracy~\cite{shaked2017improved} with stronger network and confidence predictions. 
Later, some works focused on post-process by incorporating high-level knowledge from objects such as Displets~\cite{guney2015displets}.

This inspires the study of stereo matching networks to develop a fully learnable architecture without manually designed processing. FlowNet~\cite{ilg2017flownet} are designed to find 2D optical flow by inserting two corresponding frames, which can be easily extended to stereo matching by limiting the searching within the epipolar line. PWCNet~\cite{sun2018pwc} follows a similar idea while having cost volumes calculated using a pyramid warping strategy within a local region with size of $d \times d$. However, 
for stereo estimation, one may only consider a limited range for disparity matching based on the epipolar constraint.
Therefore, to better model per-pixel disparity matching, GCNet~\cite{kendall2017end} proposes to generate a 3D cost volume of size $d \times h \times w \times c$ by densely comparing the feature at pixel (i, j) from the reference image to all possible matching pixels within the epipolar line at the target image. The network can estimate the best matching disparity through a $soft-argmin$ operation.  PSMNet~\cite{chang2018pyramid}, embracing the experience of semantic segmentation studies, which additionally exploits scale space through pyramid spatial pooling and hourglass networks to capture global image context, yielding better results than GCNet. 
As can be seen, both GCNet and PSMNet are benefited from exploring a new dimension, \ie disparity value space and scale space respectively, which motivates us to extend CSPN to 3D. Built upon PSMNet, 3D CSPN considers modeling the relationship with diffusion along their proposed new dimension, and produces more robust results.

\noindent\textbf{Spatial pyramid for hierarchical context.} As we also explore scale space for a dense prediction model, we would like to review spatial pyramid pooling (SPP)~\cite{he2014spatial} to provide more insight for our proposed model. Liu \etal~\cite{liu2015parsenet} first propose SPP to  increase the empirical receptive field of a fully convolutional network. Such an idea is demonstrated to be very effective in both semantic segmentation~\cite{chen2017rethinking,chen2016deeplab,zhao2017pyramid}, and depth estimation~\cite{chang2018pyramid}.
Here, in our perspective, the parameters for SPP form a scale space that is manually set and experimentally determined based on certain dataset by previous works~\cite{chen2017rethinking}. However, our CSPN with 3D convolution can learn the affinity for fusing the proposed scale space, which softly discovers the proper scale of context for the network. Most recently, Xu \etal~\cite{xu2018structured} also propose to learn attention guided multi-scale CRF for depth estimation, which shares similar spirits with ours.
We show in our experiments, such a strategy effectively improves the depth estimation results over PSMNet. In the near future, we will extend this idea to semantic segmentation to validate its generalization capability.

\vspace{-0.2\baselineskip}
\section{Our Approach}
\vspace{-0.6\baselineskip}
In this section, we first introduce the CSPN module we proposed, which is an anisotropic diffusion process and the diffusion tensor is learned through a deep CNN directly from the given image. Then we describe 3DCSPN, and how to use it together with SPP. Finally, we elaborate on how we applied these modules to the depth completion and stereo depth estimation.

\vspace{-1.0\baselineskip}
\subsection{Convolutional spatial propagation network (CSPN)}
\vspace{-0.\baselineskip}
\label{sec:CSPN_Module}
Given one depth map $D_o \in \textbf{R}^{m\times n}$, and one image $\ve{X} \in \textbf{R}^{m\times n}$, our task is to update the depth map to a new depth map $D_n$ within $N$ iteration steps, which not only reveals more structure details, but also improves the per-pixel depth estimation results.

\figref{fig:compare}(b) illustrates our updating operation in 2D. Formally, without loss of generality, we can embed the depth map
$D_{o} \in \textbf{R}^{m \times n}$ to some hidden space $\ve{H} \in \textbf{R}^{m \times n \times c}$, where $c$ is the number of feature channels.
The convolutional transformation functional with a kernel size of $k$ for each time step $t$ could be written as,
\begin{flalign}
    \ve{H}_{i, j, t + 1} = \kappa_{i,j}(0, 0) &\odot \ve{H}_{i, j, 0} + \sum_{\substack{a,b = -(k-1)/2\\a, b \neq 0}}^{(k-1)/2} \kappa_{i,j}(a, b) \odot \ve{H}_{i-a, j-b, t} \nonumber \\
\mbox{where,~~~~}
    \kappa_{i,j}(a, b) &= \frac{\hat{\kappa}_{i,j}(a, b)}{\sum_{a,b, a, b \neq 0} |\hat{\kappa}_{i,j}(a, b)|}, \nonumber\\
    \kappa_{i,j}(0, 0) &= \ve{1} - \sum\nolimits_{a,b, a, b \neq 0}\kappa_{i,j}(a, b)
\label{eqn:cspn}
\end{flalign}
where the transformation kernel $\hat{\kappa}_{i,j} \in \textbf{R}^{k\times k \times c}$ is the output from an affinity network, which is spatially dependent on the input image. The kernel size $k$ is usually set as an odd number so that the computational context surrounding pixel $(i, j)$ is symmetric.
$\odot$ means element-wise product here. 
Here, following SPN~\cite{liu2017learning}, we normalize kernel weights to the range of $(-1, 1)$ so that the model can be stabilized \textcolor{black}{when the condition $\sum_{a,b, a, b \neq 0} |\kappa_{i,j}(a, b)| \leq 1$ is satisfied}. 
Finally, we perform \textcolor{black}{$N$ iterations} to reach a stable status. 


\noindent\textbf{Correspondence to diffusion process with a partial differential equation (PDE).}
Similar to~\cite{liu2017learning}, here we show CSPN holds the desired property such as model stability of SPN.
Formally, we can rewrite the propagation in \equref{eqn:cspn} as a process of diffusion evolution by first doing column-first vectorization of feature map $\ve{H}$ to $\ve{H}_v \in \textbf{R}^{\by{mn}{c}}$.
\begin{align}
     &\ve{H}_v^{t+1} &= &
     \begin{bmatrix}
    0  & \kappa_{0,0}(1,0) & \cdots & 0 \\
    \kappa_{1,0}(-1,0)   & & \cdots & 0 \\
    \vdots & \vdots & \ddots & \vdots \\
    \vdots & \cdots & \cdots & 0 \\
\end{bmatrix}\ve{H}_v^{t}  + \nonumber\\
&&&\begin{bmatrix}
    1-\lambda_{0,0}  & 0 & \cdots & 0 \\
    0   & 1-\lambda_{1,0} & \cdots & 0 \\
    \vdots & \vdots & \ddots & \vdots \\
    \vdots & \cdots & \cdots & 1-\lambda_{m,n} \\
\end{bmatrix} \ve{H}_v^{0} \nonumber\\
&&=& \ve{A}\ve{H}_v^{t} + (\ve{I} - \ve{D})\ve{H}_v^{0}
\label{eqn:vector}
\end{align}
where $\lambda_{i, j} = \sum_{a,b}\kappa_{i,j}(a,b)$, and $\ve{D}$ is the degree matrix containing all the $\lambda_{i, j}$, and $\ve{A}$ is the affinity matrix. 
The diffusion process expressed as a partial differential equation (PDE) is derived as: 
\begin{align}
     \ve{H}_v^{t+1} &= (\ve{I} - \ve{D})\ve{H}_v^0 + \ve{A}\ve{H}_v^{t} \nonumber\\
     \ve{H}_v^{t+1} - \ve{H}_v^{t} &= \partial_t \ve{H}_v^{t+1} = - (\ve{I} - \ve{A}) \ve{H}_v^{t} + (\ve{I} - \ve{D})\ve{H}_v^0 
\label{eqn:proof}
\end{align}

Therefore, same as SPN~\cite{liu2017learning}, the stability of CSPN can be guaranteed if the norm of the temporal Jacobian is equal to or less than one. In our case, we follow their proof using the Gershgorin's Theorem~\cite{gershgorin1931uber}. Formally, taking \equref{eqn:proof}, for any $t$ greater than 1,  we have, 
\begin{align}
    \|\partial \ve{H}_{i, j, t + 1} / \partial \ve{H}_{i, j, t}\| = \|\ve{A}\| \leq \lambda_{max} \leq \sum_{a,b, a, b \neq 0}|\kappa_{i,j}(a, b)| \leq 1, 
\end{align}
where $\lambda_{max}$ is the maximum Eigenvalue of $\ve{A}$. This satisfies the model stability condition.

In our formulation, different from~\cite{liu2017learning} which scans the whole image in four directions sequentially~(\figref{fig:compare}(a)), CSPN propagates a local area towards all directions simultaneously at each step~(\figref{fig:compare}(b)), \ie with~\by{k}{k} local context. Larger context can be observed when recurrent processing is performed, and the context acquiring rate is in \textcolor{black}{the} order of $O(kN)$.
In \textcolor{black}{practice}, we choose to use convolutional operation due to that it can be  implemented through image vectorization, yielding efficient depth refinement.

In principal, CSPN could also be derived from mean-field approximation for solving a special type of locally connected conditional random field (CRF) model~\cite{kschischang2001factor}, where we rely the long range dependency on the propagation and the deep neural network.
However, since our approach adopts more efficient linear propagation and can be regarded as a special case of pairwise potential in graphical models, to be more accurate description, we call our strategy \textit{convolutional spatial propagation} in the field of diffusion process.

\begin{figure*}[!htpb]
	\centering
	\includegraphics[width=0.91\textwidth]{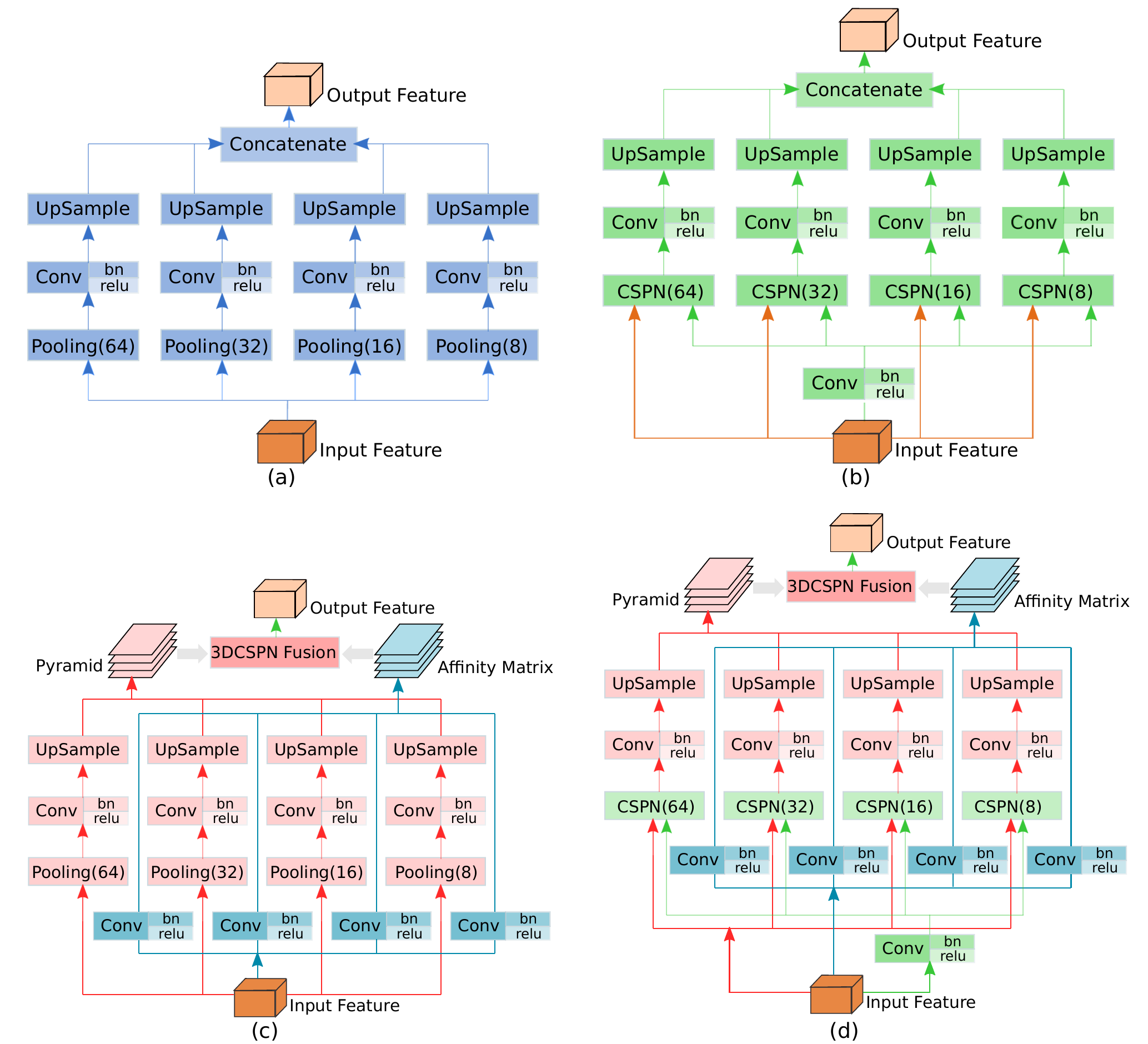}
	\caption{Different structures of context pyramid module. (a) spatial pyramid pooling (SPP) module applied by PSMNet \cite{zhao2017pyramid} (b) Our convolutional SPP (CSPP) module using 2D CSPN with different kernel size and stride. \textcolor{black}{(c)} Our convolutional feature fusion (CFF) using 3D CSPN. (d) Our final combined SPP module, namely convolutional spatial pyramid fusion (CSPF). (Details in \secref{subsec:cspn_var})}
	\label{fig:CPM}
	\vspace{-1.3\baselineskip}
\end{figure*}

\noindent\textbf{Complexity analysis.}
\label{subsec:time}
As formulated in~\equref{eqn:cspn}, our CSPN is highly parallelizable and GPU-friendly. Under the assumption of the unlimited number of computing cores for convolution, the theoretical complexity of using CUDA with GPU for one step CSPN is $O(\log_2(k^2))$, where $k$ is the kernel size. 
In theory, convolution operation can be performed \textcolor{black}{in} parallel for all pixels and channels, which has a constant complexity of $O(1)$ . Therefore, performing $N$-step propagation, the theoretical overall complexity for CSPN is $O(\log_2(k^2)N)$, which is independent of image size $(m, n)$. 

SPN~\cite{liu2017learning} adopts scanning row/column-wise propagation in four directions. Using $k$-way connection and running in parallel, the complexity for one step is $O(\log_2(k))$. The propagation needs to scan the full image from one side to another, thus the complexity for SPN is $O(\log_2(k)(m + n))$. Though this is already more efficient than the densely connected CRF~\cite{philipp2012dense}, whose implementation complexity with permutohedral lattice is $O(mnN)$, ours $O(\log_2(k^2)N)$ is more efficient since the number of iterations $N$ is always much smaller than the size of image $m, n$. 

\peng{We elaborate on the practical time cost in our experiments (\secref{sec:exp}) with a Titan X GPU. For example, with $k=3$ and $N=12$, CSPN runs 6x faster than SPN, and is $30\%$ more accurate, demonstrating both efficiency and effectiveness of the proposed approach.}

\begin{figure*}[t]
\centering
\includegraphics[width=0.9\textwidth,height=0.45\textwidth]{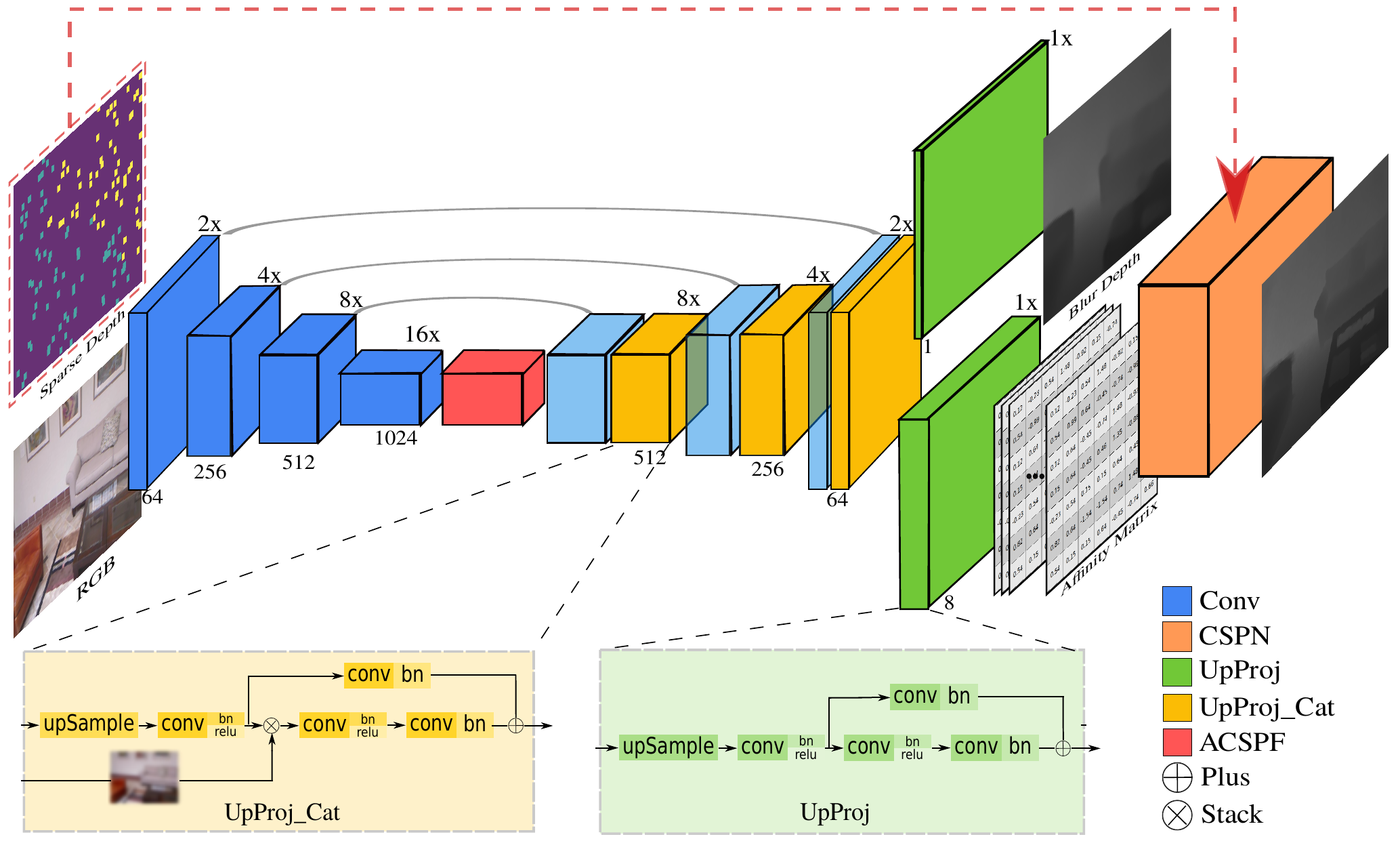}
\caption{Architecture of our networks with mirror connections for depth completion with CSPN (best view in color). Sparse depth map is embedded into the CSPN process to guide the depth refinement. The light blue blocks are the identity copy of blue blocks before. (details in \secref{subsec:unet})}
\label{fig:arch_single}
\vspace{-1.0\baselineskip}
\end{figure*}

\vspace{-.5\baselineskip}
\subsection{CSPN variants for performance boosting}
\label{subsec:cspn_var}

In this section, we present the two extensions of CSPN, \ie~3DCSPN and its conjunction with SPP.

\noindent\textbf{3DCSPN.} As introduced in \secref{sec:intro}, we extend CSPN to 3D for processing the 3D cost volume that are commonly used for stereo estimation~\cite{chang2018pyramid}. We illustrate 3DCSPN in \figref{fig:compare}(c). Similar to \equref{eqn:cspn}, given a 3D feature volume $\ve{H} \in \textbf{R}^{d \times m \times n  \times c}$, where d is an additional feature dimension, the formation for 3D CSPN could be written as, 
\begin{align}
\vspace{-1.0\baselineskip}
    \ve{H}_{i, j, l, t + 1} = \kappa_{i,j,l}(0,0,0) &\odot \ve{H}_{i, j, l, 0}  \nonumber \\
    + \sum_{\substack{a,b,c = -(k-1)/2\\a,b,c\neq 0}}^{(k-1)/2} &\kappa_{i,j,l}(a, b, c) \odot \ve{H}_{i-a, j-b, l-c, t} \nonumber \\
\mbox{where,~~~~}
    \kappa_{i,j,l}(a, b, c) &= \frac{\hat{\kappa}_{i,j,l}(a, b, c)}{\sum_{a,b,c | a, b, c \neq 0} |\hat{\kappa}_{i,j,l}(a, b, c)|}, \nonumber\\
    \kappa_{i,j,l}(0, 0, 0) &= \ve{1} - \sum\nolimits_{a,b,c | a, b, c \neq 0}\kappa_{i,j,l}(a, b, c)
\label{eqn:3dcspn}
\end{align}
which simply adds a new dimension for propagation comparing to \equref{eqn:cspn}, and we can see the original theoretical properties are all well maintained. 

\begin{figure*}[t]
\centering
\includegraphics[width=0.8\textwidth]{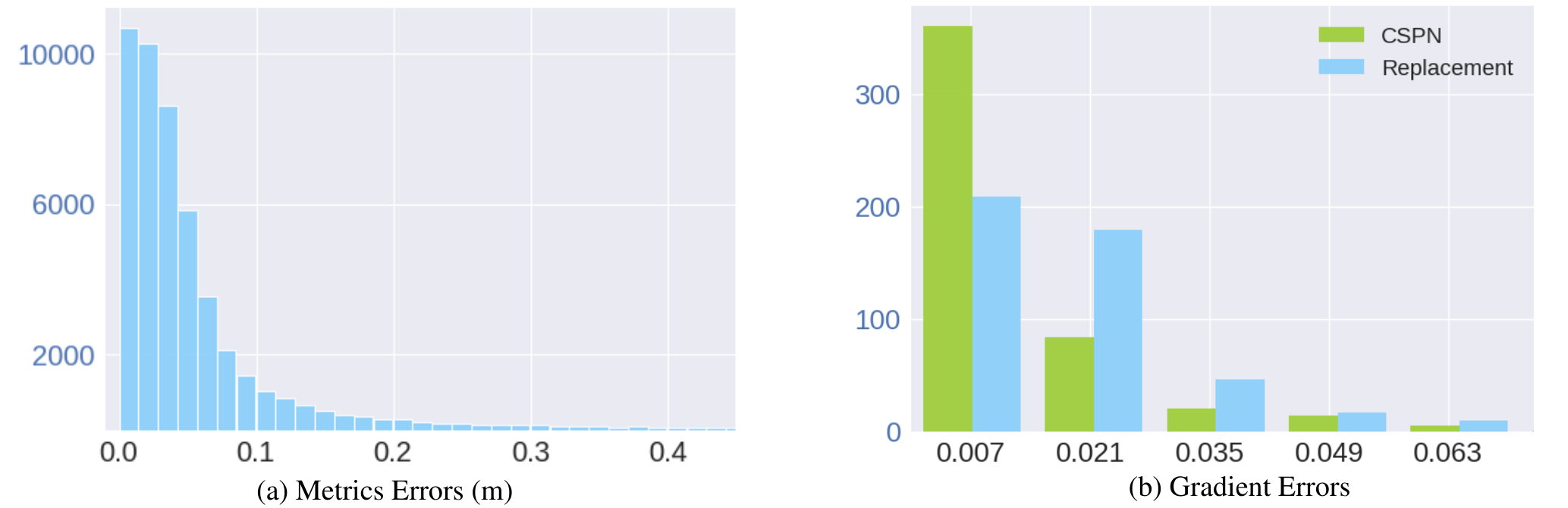}
\caption {(a) Histogram of RMSE with depth maps from Ma \etal~\cite{Ma2017SparseToDense} at given sparse depth points.  (b) Comparison of gradient error between depth maps with sparse depth replacement (blue bars) and with ours CSPN (green bars), where ours is much smaller. Check~\figref{fig:gradient} for an example. Vertical axis shows the count of pixels.}
\label{fig:hist}
\end{figure*}
\begin{figure*}[t]
\centering
\includegraphics[width=0.9\textwidth]{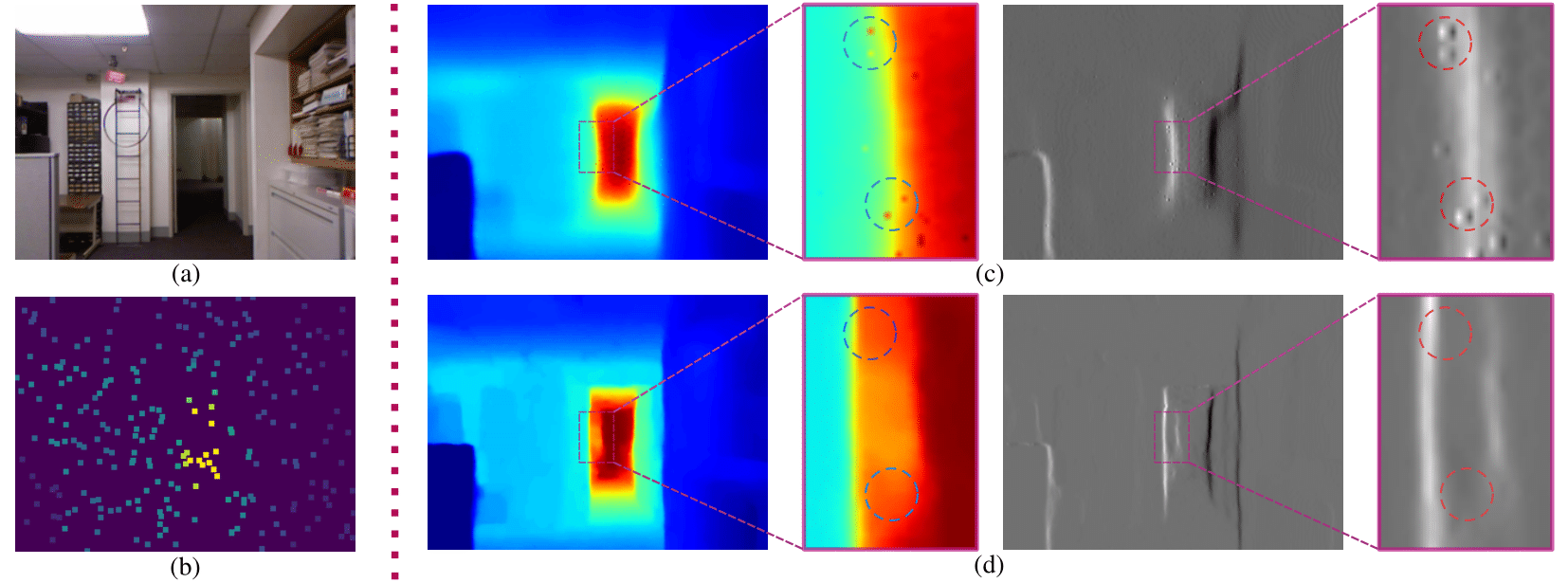}
\caption{Comparison of depth map~\cite{Ma2017SparseToDense} with sparse depth replacement and with our CSPN \wrt smoothness of depth gradient at sparse depth points. (a) Input image. (b) Sparse depth points. (c) Depth map with sparse depth replacement. \textit{Left}: Depth map. \textit{Right}: Sobel gradient in the x-axis direction (d) Depth map with our CSPN with sparse depth points. We highlight the differences in the red box.}
\label{fig:gradient}
\vspace{-1.0\baselineskip}
\end{figure*}

\noindent\textbf{Convolutional spatial pyramid fusion (CSPF).}
The second module in the architecture we enhanced is the spatial pyramid pooling (SPP) as illustrated in \figref{fig:CPM}(a). 
Here, we can see that each branch of SPP can be treated as a special case of one-step CSPN given proper kernel size and convolution stride. Formally, given a feature map with the size of $h \times w \times c$, and target pooled feature map with spatial size of $p \times q$, spatial pooling computes the mean value within each parted grid with size of $h/p \times w/q$. This is equivalent to one step CSPN (\equref{eqn:cspn}) by setting both convolution kernel size and stride to be $h/p \times w/q$, and all the values in $\kappa(a, b)$ to be uniform. 
However, we know that features can be very different at impacting the final performance as shown in many attention models~\cite{vaswani2017attention}.
Therefore, we propose to learn such a pooling/transformation kernel $\kappa(a, b)$ using CSPN for this SPP module. As shown in \figref{fig:CPM}(b), in our case, we output an affinity matrix from the same feature block for spatial pooling, based on which we do one step 2D CSPN, yielding the required pooled feature map with size of $p \times q$. 
Specifically, feature maps with target sizes of 64 x 64, 32 x 32, 16 x16 and 8 x 8 are adopted (\figref{fig:CPM}(a)), and all the feature maps share the same network output for computing the pooling kernels. 
In practice, our network outputs a one channel weight map with size of $h \times w \times 1$, and for each target size of pooled feature, we first partitioning the weight map to pooling regions, and compute the pooling/transformation kernel $\kappa()$ within each region. However, rather than using the normalization in \equref{eqn:cspn}, we enforce the kernel value to be positive, and normalize the them over all output values without dropping the kernel value of center pixel, which is formally written as, $\kappa_{i,j}(a, b) = \frac{|\hat{\kappa}_{i,j}(a, b)|}{\sum_{a,b} |\hat{\kappa}_{i,j}(a, b)|}$. Here, the output weight map is repeated $c$ times to match the feature dimension in CSPN. We call our strategy of multi-scale feature computation as \textit{convolutional spatial pyramid pooling} (CSPP) to simplify our description later.

Last, we need to fuse the $l$ feature maps from all the layers of the spatial pyramid. Rather than \textcolor{black}{directly} concatenating all the pooled features into a feature map with size $ h \times w \times lc$ as usual~\cite{chang2018pyramid}, we adopt the strategy of weighted averaging, which is illustrated in \figref{fig:CPM}(c). Specifically, we concatenate the output spatial pyramid features into a 4D feature volume with size $l \times h \times w  \times c$, and learn a transformation kernel with size of $l \times 3 \times 3 $. After doing one step 3D CSPN without padding in the layer dimension, \ie use padding size of $[0, 1, 1]$, we obtain a fused feature map with size $h \times w \times c$. Here, we also adopt the same kernel normalization as that in CSPP. 
Here, we use one independent branch for computing the transformation kernel for each layer of spatial pyramid, and concatenate them to a 3D CSPN kernel.
We call this strategy as \textit{convolutional feature fusion} (CFF) to simplify our description. 
As shown in \figref{fig:CPM}(d), our final spatial pooling strategy is a combination of CSPP and CFF, which we call convolutional spatial pyramid fusion (CSPF). CSPF produces significant performance boost over the original SPP module in our tasks, especially for stereo depth estimation as demonstrated in \secref{subsec:stereo_exp}.

Finally, we also adopt Atrous SPP (ASPP)~\cite{ChenPSA17} to replace SPP for multi-scale feature pooling without feature size reduction. Specifically, ASPP use dilated convolution to obtain features within various context. Similarly, our CSPN can be performed in the same manner of dilated convolution by learning a spatial dependent transformation kernel for each layer of the pyramid. Therefore, we also extend ASPP to atrous CSPP (ACSPP) for computing spatial pyramid features. In our experiments, we use the set of dilation rates proposed in ASPP~\cite{ChenPSA17} including 6 x 6, 12 x 12, 18 x18 and 24 x 24, and found ACSPP achieves better performance than that from CSPP. Combining it with CFF, called ACSPF, yields our best setting for competing over the benchmarks.


\subsection{Learning depth completion with CSPN}
In depth completion, we have a sparse depth map $D_s$ (\figref{fig:arch_single}) joint with an RGB image as inputs to our model. Specifically, a sparse set of pixels already have known depth values (obtained by other sensors). These pixels are used to guide the depth estimation for the remaining pixels. As shown in \figref{fig:arch_single}, our architecture is similar to that from Ma \etal~\cite{Ma2017SparseToDense}, but with three major modifications: a ACSPF module (red block), a CSPN layer with sparse depths (orange block) and mirror connections. Later, we will elaborate the motivation and design of the second and third modifications.


\subsubsection{Spatial propagation with sparse depths.}
\label{subsec:sparse2dense}
As stated in \secref{sec:CSPN_Module}, CSPN help to recover better details. However, for depth completion, we need to preserve the depth values at those valid pixels with known depth. To achieve this goal, we modify CSPN to include the sparse depth map $D_s$ in the diffusion process. Specifically, we embed $D_s$ to a hidden representation $\ve{H}^s$, and re-write the \textcolor{black}{update} equation of $\ve{H}$ by adding a replacement step after performing \equref{eqn:cspn}, 
\begin{align}
    \ve{H}_{i, j, t+1} = (1 - m_{i, j}) \ve{H}_{i, j, t+1}  +  m_{i, j} \ve{H}_{i, j}^s 
\label{eqn:cspn_sp}
\end{align}
where $m_{i, j} = \textbf{I}(d_{i, j}^s > 0)$ is an indicator for the validity of sparse depth at $(i, j)$.

\begin{figure*}[t]
	\centering
	\includegraphics[width=1.0\textwidth]{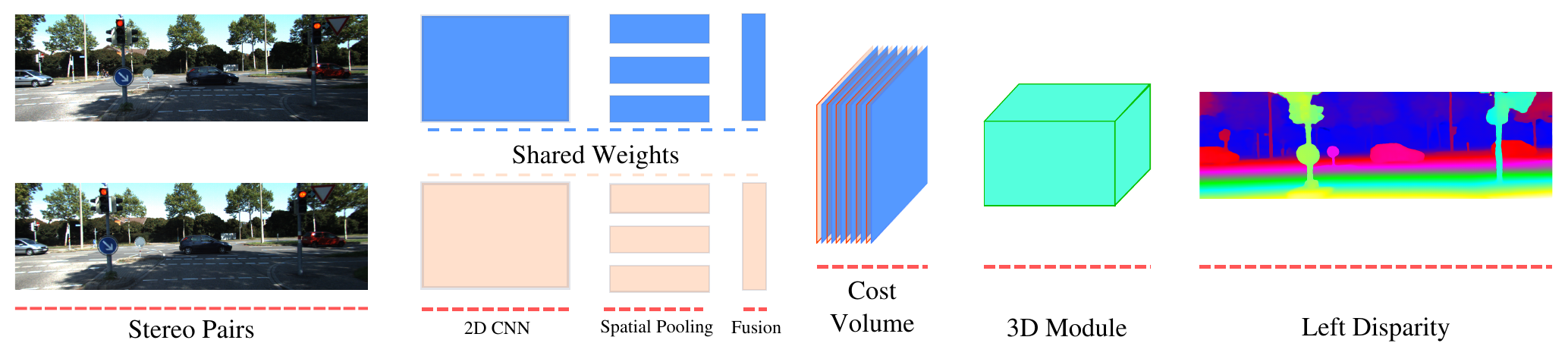}
	\caption{Architecture of our networks for stereo depth estimation via transformation kernel prediction with 3D CSPN (best view in color).}
	\label{fig:arch_all}
\end{figure*}

\begin{figure*}[t]
	\centering
	\includegraphics[width=0.9\textwidth]{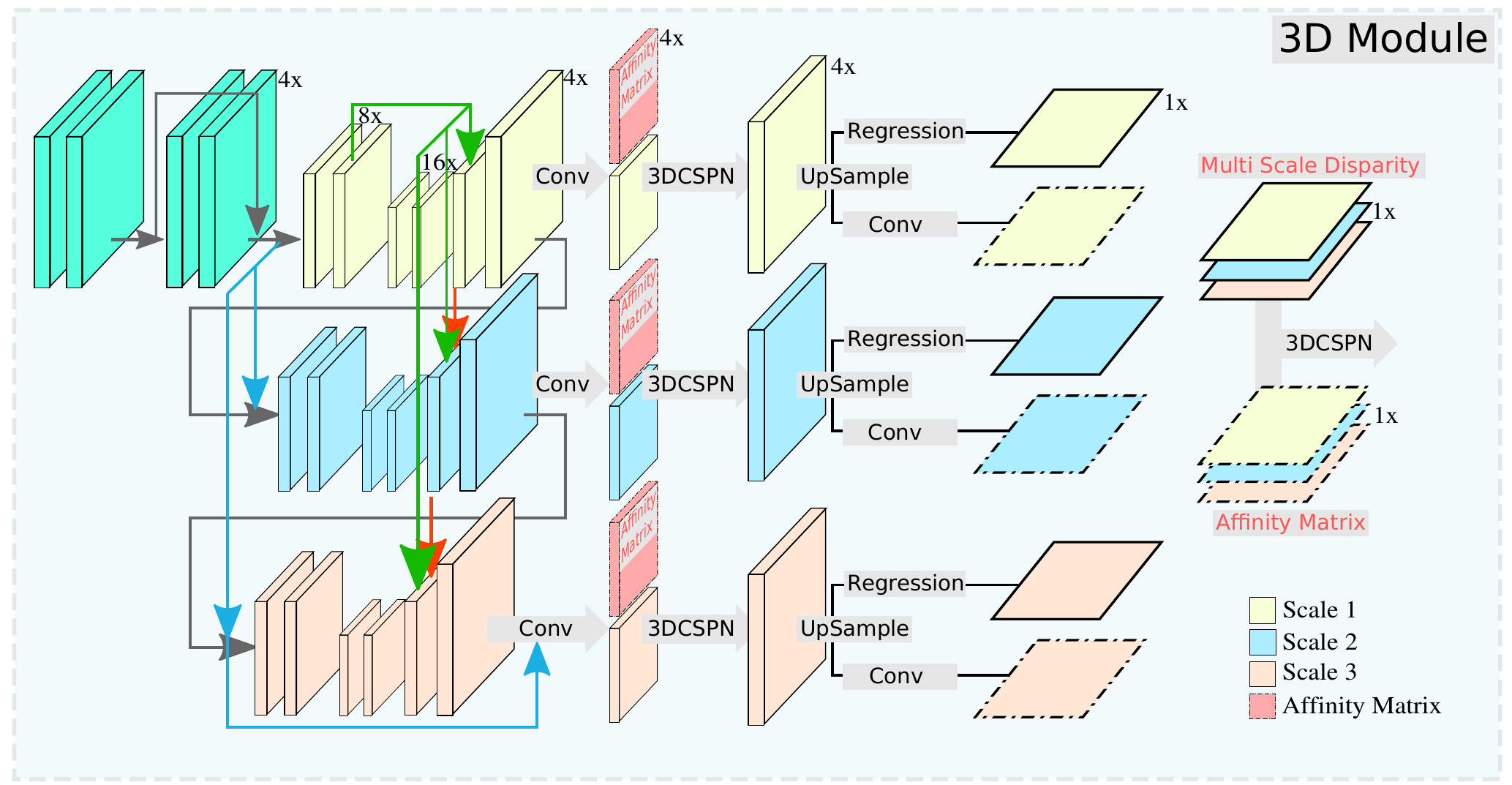}
	\caption{ Details of our 3D Module (best view in color). Downsample rate \wrt image size is shown at the right top corner of each block, \eg~4x means the size of the feature map is $\frac{h}{4} \times \frac{w}{4}$ where $h \times w$ is image size. \peng{The red, green and blue arrows are skip connections, indicating feature concatenation at particular position, which are the same with PSMNet~\cite{zhao2017pyramid}.}}
	\label{fig:arch}
\end{figure*}

From \equref{eqn:cspn_sp}, the updating still follows the diffusion process with PDE, where the affinity matrix can be built by simply replacing the rows satisfying $m_{i, j} = 1$ in $\ve{G}$ (\equref{eqn:vector}) with $\ve{e}_{i + j*m}^T$. Here $\ve{e}_{i + j*m}$ is an unit vector with the value at $i + j*m$ as 1.
Therefore, the summation of each row is still $1$, and the stabilization still holds in this case.

With CSPN, we propagate the information from those sparse depth to its surrounding pixels such that the smoothness between the sparse depths and their neighbors are maintained, and thanks to the learned image dependent affinity, the final depth map is well aligned with image structures.


Our strategy has several advantages over the previous state-of-the-art sparse-to-dense methods~\cite{Ma2017SparseToDense,LiaoHWKYL16,ma2018self}.
In \figref{fig:hist}(a), we plot a histogram of depth displacement from ground truth at given sparse depth pixels from the output of Ma \etal~\cite{Ma2017SparseToDense}. It shows that the original values of sparse depth points cannot \textcolor{black}{be} preserved, and some pixels could have very large displacement (0.2m), indicating that directly training a CNN for depth prediction does not preserve the value of real sparse depths provided. To ensure such a property, 
one may simply replace the depths from the outputs withthe original values at these pixels with known depth, however, this yields non-smooth depth gradient \wrt surrounding pixels. 
In~\figref{fig:gradient}(c), we plot such an example, at right of the figure, we compute Sobel gradient~\cite{sobel2014history} of the depth map along x direction, where we can clearly see that the gradients surrounding pixels with replaced depth values are non-smooth.
We statistically verify this in \figref{fig:hist}(b) using 500 sparse samples, the blue bars are the histogram of gradient error  at sparse pixels by comparing the gradient of the depth map with sparse depth replacement and of ground truth depth map. We can see the difference is significant, 2/3 of the sparse pixels has large gradient error.
Our method, on the other hand, as shown with the green bars in  \figref{fig:hist}(b), the average gradient error is much smaller, and most pixels have no error at all. In~\figref{fig:gradient}(d), we show the depth gradients surrounding sparse pixels are smooth and close to ground truth, demonstrating the effectiveness of our propagation scheme.

\subsubsection{Mirror connections for better details}
\label{subsec:unet}

In CSPN, in order to recover spatial details in the input image, the network for learning affinity matrix should contain less spatial down-sampling. Therefore, Liu \etal~\cite{liu2017learning} build a separate deep network for affinity learning. In our case, we wish to share the feature extractors for affinity and depth estimation, which saves us both memory and time cost for learning and inference. Therefore, as shown in \figref{fig:arch_single},
we fork an additional output from the given depth network to predict the affinity matrix.
 
 However, spatial information is \textcolor{black}{weakened} or even lost with the down sampling operation during the forward process of the ResNet in~\cite{laina2016deeper}. Thus, we add mirror connections similar with the U-shape network~\cite{ronneberger2015u} by \textcolor{black}{directly} concatenating the feature from encoder to up-projection layers as illustrated by ``UpProj$\_$Cat'' layer in~\figref{fig:arch_single}. 
 Notice that it is important to carefully select the end-point of mirror connections. Through experimenting three possible positions to append the connection, \ie after \textit{conv}, after \textit{bn} and after \textit{relu} as shown by the ``UpProj'' layer in~\figref{fig:arch_single} , we found the last position provides the best results in the NYU v2 dataset (\secref{subsec:ablation}). 
In this configuration, we found not only the depth output from the network is better recovered, but also the results after CSPN is further refined.
Finally we adopt the same training loss as~\cite{Ma2017SparseToDense}, yielding an end-to-end learning system.

\subsection{Learning stereo matching with CSPN}
\label{subsec:stereo}

In this application, we adopt a network architecture similar to PSMNet~\cite{zhao2017pyramid} as shown in \figref{fig:arch_all}. The left and right images are fed to two weight-sharing CNN, yielding corresponding feature maps, a spatial pooling module for feature harvesting by concatenating representations from sub-regions with different sizes. The two produced feature maps are then used to form a 4D cost volume, which is fed into a 3D CNN for disparity regression. 
We refer readers to the original paper for more details due to space limitation. 
Here, we made two changes by replacing CSPF with their spatial pooling module and appending 3DCSPN after multi-scale outputs. 
Here, we update the spatial pooling and 3D module with our proposed CSPF and 3DCSPN (\secref{subsec:cspn_var}), it is straight-forward to use CSPF and we describe the details of applying 3DCSPN in the following.


\subsubsection{3DCSPN over disparity and scale space}
\label{subsubsec:3d_cspn}
In \figref{fig:arch}, we zoom into the 3D module of \figref{fig:arch_all} to clarify the 3D CSPN we applied for disparity regression. In PSMNet, three predicted disparity volumes with size of $d/4 \times h/4 \times w/4 \times 1$ are output at different stages from a stacked hourglass network. Here $d, h, w$ is the maximum disparity, height and width of the input image correspondingly. 
Similar to the appending strategy of 2D CSPN for single image depth prediction in~\secref{subsec:unet}, after the disparity volume at each stage, we append a 3D CSPN with kernel size $k \times k \times k$ to combine the contexts from neighbor pixels, where the affinity matrix is learned from the same feature block as the outputs. Then,  \textcolor{black}{trilinear} upsampling is applied to upsample a disparity volume to $d \times h \times w \times 1$ for disparity map regression, yielding an output with shape of $h \times w \times 1$. Here, 3DCSPN finish processing over the disparity space (ds). 

To fuse the multiple disparity maps output from different stages, PSMNet manually sets the weight to average the outputs. In our case, we concatenate them into a 4D volume with size $s \times h \times w \times 1$, where $s=3$ is the number of disparity maps. 
Similarly, we can perform a 3D CSPN with kernel size as $s \times k \times k$ to connect the multi-stage predictions, which is conceptually similar as attention models for multi-scale feature fusion~\cite{Chen_2016_CVPR}. Last, we use feature padding with size of $[0, 1, 1]$, so that the first dimension is reduced to $1$ with one iteration, and we obtain a single regressed disparity map with shape $h \times w \times 1$ for final depth estimation. Here, 3DCSPN finish processing over the scale space (ss). 




Finally, for training the full network, We use the same $soft$-$argmin$ disparity regression method proposed by GCNet~\cite{kendall2017end} to convert final discretized disparity values to continuous ones.
\begin{equation}
\widehat{d}=\sum_{d=0}^{D_{max}} d \cdot \sigma \left ( -c_{d} \right )
\label{eqn:disp}
\end{equation}

Then, the continuous disparity value is compared against the ground truth disparity value using the $L_{1}$ loss. Formally, the loss function is defined as:
\begin{equation}
\label{func:loss}
L\left (d^*, \widehat{d}~\right ) = \frac{1}{N}\sum_{i=1}^{N} \|d^* - \widehat{d}\|_1,
\end{equation}
where  $d^*$ is a ground truth disparity, and $\widehat{d}$ is the predicted disparity from \equref{eqn:disp}.

\begin{figure*}[t]
\centering
\includegraphics[width=1.0\textwidth]{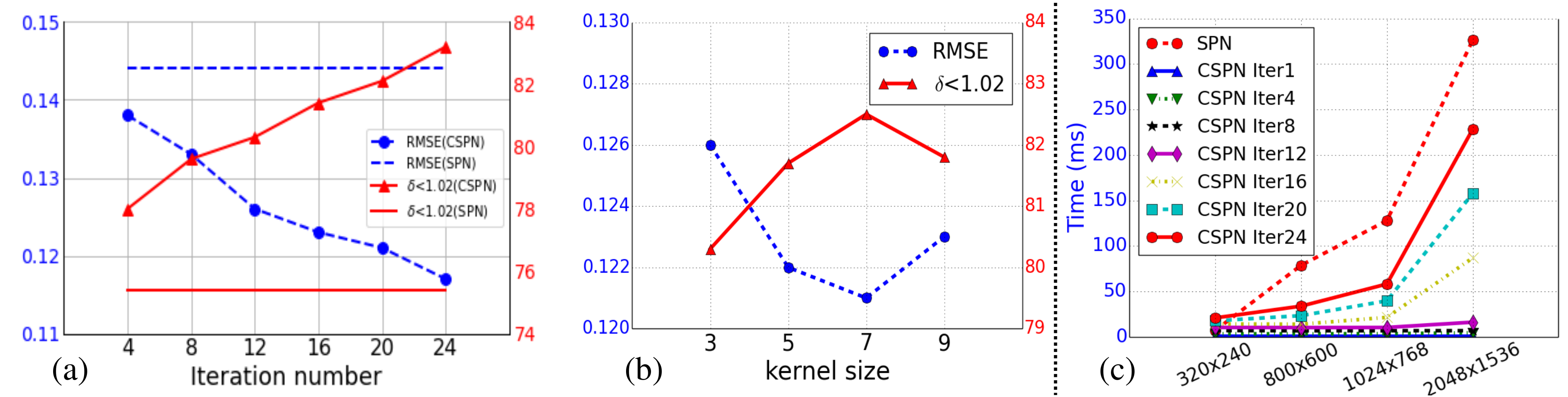}
\caption{Ablation study.(a) RMSE (left axis, lower the better) and $\delta < 1.02$ (right axis, higher the better) of CSPN \wrt number of iterations. Horizontal lines show the corresponding results from SPN~\cite{liu2017learning}. (b) RMSE and $\delta < 1.02$ of CSPN \wrt kernel size. (c) Testing times \wrt input image size.}
\label{fig:ab_study}
\end{figure*}

\section{Experiments}
\label{sec:exp}
In this section, we first describe our implementation details, the datasets and evaluation metrics used in our experiments, then present a comprehensive evaluation of CSPN and its variants, \ie~CSPF and 3D CSPN, on the two tasks of depth completion and stereo matching.

\begin{table*}[t]
	\centering
	\fontsize{8.5}{8.5}\selectfont
	\caption{Comparison results on NYU v2 dataset~\cite{silberman2012indoor} between different variants of CSPN and other state-of-the-art strategies. Here, ``w'' means adding corresponding components inline to the SOTA baseline architecture~\cite{Ma2017SparseToDense}. ``MC'' is short for mirror connection. ``Preserve SD'' is short for preserving the depth value at sparse depth samples.}
	\bgroup
	\def\arraystretch{1.3}
	\setlength{\tabcolsep}{6pt} 
	\begin{tabular}{lccc|cccccc}
		\hline
		\multicolumn{1}{l}{\multirow{2}{*}{Method}}  & \multirow{2}{*}{~~Preserve ``SD''~~} & \multicolumn{2}{c|}{~~Lower the better~~}   & \multicolumn{6}{c}{Higher the better} \\ \cline{3-10}
		\multicolumn{1}{l}{} & & ~~RMSE~~  & \multicolumn{1}{c|}{~~REL~~} & ~~$\delta_{1.02}$~~ & ~~$\delta_{1.05}$~~ & ~~$\delta_{1.10}$~~ & ~~$\delta_{1.25}$~~ & ~~$\delta_{1.25^2}$~~ & ~~$\delta_{1.25^3}$~~ \\ \hline
		~~Ma \etal \cite{Ma2017SparseToDense} &   & 0.230  & 0.044    & 52.3            & 82.3            & 92.6            & 97.1            & 99.4              & 99.8              \\ \hline 
		w Bilateral~\cite{barron2016fast}&   & 0.479  & 0.084    & 29.9            & 58.0            & 77.3            & 92.4            & 97.6              & 98.9     \\\hline
		w DenseCRF~\cite{philipp2012dense}&   & 0.177  &0.032     & 51.9            & 84.5            & 94.0            & 98.4            & 99.6              & 99.9     \\\hline
		w SPN~\cite{liu2016learning}      &  & 0.172          & 0.031                    & 61.1            & 84.9            & 93.5            & 98.3            & 99.7              & 99.9              \\ \hline
	    w 2D CSPN (Ours)     &  & 0.162          & 0.028                    & 64.6            & 87.7            & 94.9            & 98.6            & 99.7              & 99.9              \\\hline
	    w Mirror connection (MC) (Ours)     &  & 0.137          & 0.020                    & 78.1       & 91.6 & 96.2 & 98.9 & 99.8 & 100.0              \\ \hline
        \hline
        w Replacement &\checkmark   & 0.168  & 0.032    & 56.5            & 85.7            & 94.4            & 98.4            & 99.7              & 99.8              \\ \hline
        w ARAP~\cite{igarashi2005rigid} & \checkmark  & 0.232  & 0.037    & 59.7            & 82.5            & 91.3            & 97.0            & 99.2              & 99.7     \\\hline
        w SPN~\cite{liu2016learning} &\checkmark & 0.162    & 0.027      & 67.5            & 87.9            & 94.7            & 98.5            & 99.7              & 99.9  \\\hline
        w CSPN   & \checkmark & 0.136    & 0.021      & 76.2  & 91.2  & 96.2   & 99.0   &99.8     & 100.0   \\ \hline
        w MC+SPN~\cite{liu2016learning}      & \checkmark &  0.144 & 0.022 & 75.4 & 90.8 & 95.8 & 98.8 & 99.8 & 100.0   \\ \hline
        w MC+CSPN (Ours) & \checkmark & 0.117 & 0.016 & 83.4 & 93.5 & 97.1 & 99.2 & 99.9 & 100.0   \\ \hline
        w MC+CSPN+ASPP (Ours) & \checkmark & 0.116 & 0.016 & 83.6 & 93.5 & 97.1 & 99.2 & 99.9 & 100.0   \\ \hline       
        w MC+CSPN+ASPP+CFF (Ours) & \checkmark & 0.114 & 0.015 & 84.3 & 93.8 & 97.2 & 99.3 & 99.9 & 100.0   \\ \hline
        w MC+CSPN+ACSPP (Ours) & \checkmark & 0.113 & 0.015 & 83.8 & 93.7 & 97.2 & 99.3 & 99.9 & 100.0   \\ \hline
        w MC+CSPN+ACSPF (Ours) & \checkmark & \textbf{0.112} & \textbf{0.015} & \textbf{84.7} & \textbf{93.9} & \textbf{97.3} & \textbf{99.3} & \textbf{99.9} & \textbf{100.0}   \\ \hline
	\end{tabular}
	\egroup
\label{tbl:sota}
\vspace{-1.0\baselineskip}
\end{table*}

\begin{figure*}[t]
\centering
\includegraphics[width=0.98\textwidth]{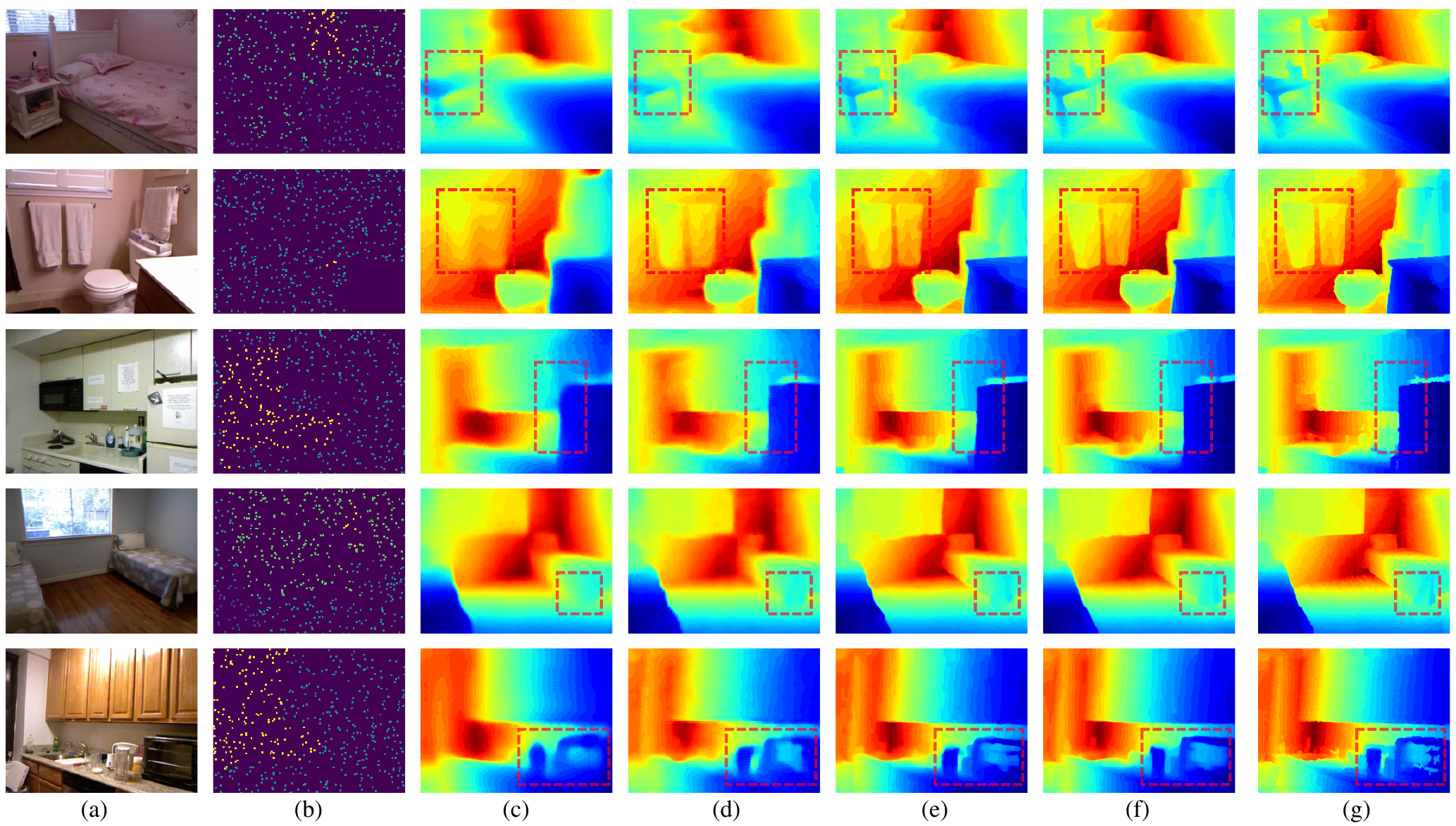}
\caption{Qualitative comparisons on NYU v2 dataset. (a) Input image; (b) Sparse depth samples(500); (c) Ma \etal \cite{Ma2017SparseToDense}; (d) Mirror connection (MC)+SPN\cite{liu2016learning}; (e) MC+CSPN(Ours); (f) MC+CSPN+CSPF (Ours);(g) Ground Truth. Most significantly improved regions are highlighted with dash boxes (best view in color).}
\label{fig:nyudepth}
\vspace{-1.3\baselineskip}
\end{figure*}

\subsection{Depth completion}
\label{subsec:depthcomp}

\noindent\textbf{Implementation details.}
Following the network proposed in~\cite{laina2016deeper,Ma2017SparseToDense}, the weights of ResNet in the encoding layers for depth estimation (\secref{subsec:unet}) are initialized with models pretrained on the ImageNet dataset~\cite{deng2009imagenet}.  
Our models are trained with SGD optimizer, and we use a small batch size of 8 and train for 40 epochs for all the experiments, and the model performed best on the validation set is used for testing. The learning rate starts at 0.01 ,and will reduced to 20$\%$ when there is not any improvements in three consecutive epochs. 
A small weight decay of $10^{-4}$ is applied for regularization.

For depth, we found out that propagation with hidden representation $\ve{H}$ achieved marginal improvement over doing propagation within the domain of depth $D$. Therefore, we perform all our experiments directly with $D$ rather than learning an additional embedding layer. For sparse depth samples, we adopt 500 sparse samples as that is used in~\cite{Ma2017SparseToDense}.

\subsubsection{Datasets and Metrics}
\label{subsec:data_metric}
All our experiments are evaluated on two datasets: NYU v2~\cite{silberman2012indoor} and KITTI Odometry ~\cite{geiger2012we}, using commonly used metrics.

\noindent\textbf{NYU v2.} The NYU-Depth-v2 dataset consists of RGB and depth images collected from 464 different indoor scenes. We use the official split of data, where 249 scenes are used for training and we sample 50K images out of the training set with the same manner as~\cite{Ma2017SparseToDense}. For testing, following the standard setting~\cite{eigen2015predicting,peng2016depth}, the small labeled test set with 654 images is used the final performance. The original image of size $\by{640}{480}$ are first downsampled to half and then center-cropped, producing a network input size of $\by{304}{228}$.

\noindent\textbf{KITTI odometry dataset.} It includes both camera and LiDAR measurements, and consists of 22 sequences. Half of the sequence is
used for training while the other half is for evaluation. Following~\cite{Ma2017SparseToDense}, we use all 46k images from the training sequences for training, and a random subset of 3200 images from the test sequences for evaluation. Specifically, we take the bottom part $\by{912}{228}$ due to the lack of depth at the top area, and only evaluate the pixels with ground truth.

\noindent\textbf{Metrics.} We adopt the same metrics and use their implementation in \cite{Ma2017SparseToDense}. Given ground truth depth $D^* = \{d^*\}$ and predicted depth $D = \{d\}$, the metrics include:  (1) RMSE: $\sqrt{\frac{1}{|D|}\sum_{d \in D}||d^* - d||^2}$. (2) Abs Rel: $\frac{1}{|D|}\sum_{d \in D}|d^* - d|/d^*$. (3) $\delta_t$: $\%$ of $d \in D$, s.t. $max(\frac{d^*}{d}, \frac{d}{d^*})<t$, where $t \in \{1.25, 1.25^2, 1.25^3\}$. Nevertheless, for the third metric, we found that the depth accuracy is very high when sparse depth is provided, $t = 1.25$ is already a very loose criteria where  almost $100\%$ of pixels are judged as correct, which can hardly distinguish different methods as shown in (\tabref{tbl:sota}). Thus we adopt more strict criteria for correctness by choosing $t \in \{1.02, 1.05, 1.10\}$.





\subsubsection{Ablation study for CSPN Module}
\label{subsec:ablation}
Here, we evaluate various hyper-parameters including kernel size $k$, number of iterations $N$ in \equref{eqn:cspn} using the NYU v2 dataset for single image depth estimation. Then we provide an empirical evaluation of the running speed with a Titan X GPU on a computer with 16 GB memory.

\noindent\textbf{Number of iterations.} We adopt a kernel size of $3$ to validate the effect of iteration number $N$ in CSPN. 
As shown in  \figref{fig:ab_study}(a), our CSPN has outperformed SPN~\cite{liu2017learning} (horizontal line) when iterated only four times. Also, we can get even better performance when more iterations are applied in the model during training. From our experiments, the accuracy is saturated when the number of iterations is increased to $24$, which we adopt for final evaluation. 

\noindent\textbf{Size of convolutional kernel.} As shown in  \figref{fig:ab_study}(b), larger convolutional kernel has similar effect with more iterations, due to the larger context being considered for propagation at each time step. Here, we hold the iteration number to $N = 12$, and we can see the performance is better when $k$ is larger while saturated at size of $7$. 
We notice that the performance drops slightly when the kernel size is set to $9$. This is because that we use a fixed number of epoch, \ie 40, for all the experiments, while larger kernel size induces more affinity to learn in propagation, which needs more epoch of data to converge. Later, when we train with more epochs, the model reaches similar performance with kernel size of $7$.
Thus, we can see using kernel size of $7$ with $12$ iterations reaches similar performance of using kernel size of $3$ with $20$ iterations, which shows CSPN has the trade-off between kernel size and iterations. In practice, the two settings run with similar speed, while the latter costs much less memory. Therefore, we adopt kernel size as $3$ and number of iterations as $24$ in our comparisons.

\noindent\textbf{Concatenation end-point for mirror connection.} 
As discussed in \secref{subsec:unet}, based on the given metrics, we experimented three concatenation places, \ie after \textit{conv}, after \textit{bn} and after \textit{relu} by fine-tuning with weights initialized from encoder network trained without mirror-connections.
The corresponding RMSE are  $0.531$, $0.158$ and $0.137$ correspondingly. Therefore, we adopt the proposed concatenation end-point.

\noindent\textbf{Running speed}
For testing, in  \figref{fig:ab_study}(c), we show the running time comparison between the SPN and CSPN with kernel size as $3$. We use the author's PyTorch implementation online. As can be seen, we can get better performance within much less time for large images, while is comparable when image size is small $320 \times 240$. For example, four iterations of CSPN on one $1024\times768$ image only takes 3.689~$ms$, while SPN takes 127.902~$ms$. In addition, the time cost of SPN is linearly growing \wrt image size, while the time cost of CSPN is independent to image size and much faster as analyzed in \secref{subsec:time}. In practice, however, when the number of iterations is large, \eg ``CSPN Iter 20'', we found the practical time cost of CSPN also grows \wrt image size. 
In principle, we can eliminate such a memory bottleneck by customizing a new operation, which will be our future work. Nevertheless, without coding optimation, even at high iterations with large images, CSPN's speed is still twice as fast as SPN. 

\begin{table}[!htpb]
    \centering
    \begin{tabular}{ l | c c  c }
    
        \multirow{2}{*}{Method} & Memory & Inference & Train  \\
        & (MB/batch) &  (ms/image) & (s/batch) \\
        \hline
        Ma \etal \cite{Ma2017SparseToDense} &4610  & 4.99  & 0.158 \\
        w MC  & 9154 & 7.73 & 0.172 \\
        w MC+SPN~\cite{liu2016learning} & 11158 & 11.73 & 0.189 \\
        w MC+CSPN  &  11526 & 10.77  & 0.184 \\
        w MC+CSPN+ACSPF & 11824 & 14.56 & 0.195 \\\hline
    \end{tabular}
    \caption{Average memory cost, training and inference time on NYU v2 dataset of various models with batch size as 8 and image size as 304$\times$ 228 (4 iterations in CSPN). }
    \label{tab:time}
    \vspace{-1.0\baselineskip}
\end{table}

\peng{\tabref{tab:time} shows training and inference time in NYU v2.  Our model memory cost for the single batch of NYU v2 is about 12GB, which is larger than Ma \etal~\cite{Ma2017SparseToDense} and comparable to SPN~\cite{liu2016learning}. The learning/inference time for each epoch with 50K images is 0.9 hours, which is comparable to SPN~\cite{liu2016learning}. }

\begin{table*}[!htpb]
	\centering
	\caption{Comparison results on KITTI dataset~\cite{geiger2012we} 
	}
	\label{tbl:sota_kitti}
	\fontsize{8.5}{9.0}\selectfont
	\bgroup
	\def\arraystretch{1.3}
	\setlength{\tabcolsep}{6pt} 
    \begin{tabular}{lccc|cccccc}
	\hline
		\multicolumn{1}{l}{\multirow{2}{*}{Method}}  & \multirow{2}{*}{~~Preserve ``SD''~~} & \multicolumn{2}{c|}{~~Lower the better~~}   & \multicolumn{6}{c}{Higher the better} \\ \cline{3-10}
		\multicolumn{1}{l}{} & & ~~RMSE~~  & \multicolumn{1}{c|}{~~REL~~} & ~~$\delta_{1.02}$~~ & ~~$\delta_{1.05}$~~ & ~~$\delta_{1.10}$~~ & ~~$\delta_{1.25}$~~ & ~~$\delta_{1.25^2}$~~ & ~~$\delta_{1.25^3}$~~ \\ \hline
	~~Ma \etal ~\cite{Ma2017SparseToDense}                    &                                 & 3.378                                 & 0.073                    & 30.0            & 65.8            & 85.2            & 93.5            & 97.6              & 98.9              \\ \hline
	w SPN~\cite{liu2016learning}                   & \checkmark                                & 3.243                                 & 0.063                    & 37.6            & 74.8            & 86.0            & 94.3            & 97.8              & 99.1              \\ \hline
	w Mirror connection (MC) (Ours)            &                              & 3.049                                 & 0.051                    & 62.6            & 83.2            & 90.2            & 95.3            & 97.9              & 99.0              \\ \hline
	w CSPN(Ours)            & \checkmark                                & 3.029                                 & 0.049                    & 66.6            & 83.9            & 90.7            & 95.5            & 98.0              & 99.0              \\ \hline
	w MC+SPN        & \checkmark                               & 3.248                                 & 0.059                    & 52.1            & 79.0            & 87.9            & 94.4            & 97.7              & 98.9              \\ \hline
	w MC+CSPN(Ours)       & \checkmark                                & 2.977                        & 0.044           & 70.2   & 85.7   & 91.4   & 95.7   & 98.0     & 99.1     \\ \hline
	w  MC+CSPN+ACSPF (Ours) & \checkmark                                & \textbf{2.843}                        & \textbf{0.042}           & \textbf{72.9}   & \textbf{86.6}   & \textbf{92.2}   & \textbf{96.1}   & \textbf{98.2}     & \textbf{99.2}     \\ \hline
    \end{tabular}
	\egroup
\label{tbl:sota_kitti}
\end{table*}

\begin{figure*}[!htpb]
\centering
\includegraphics[width=0.95\textwidth]{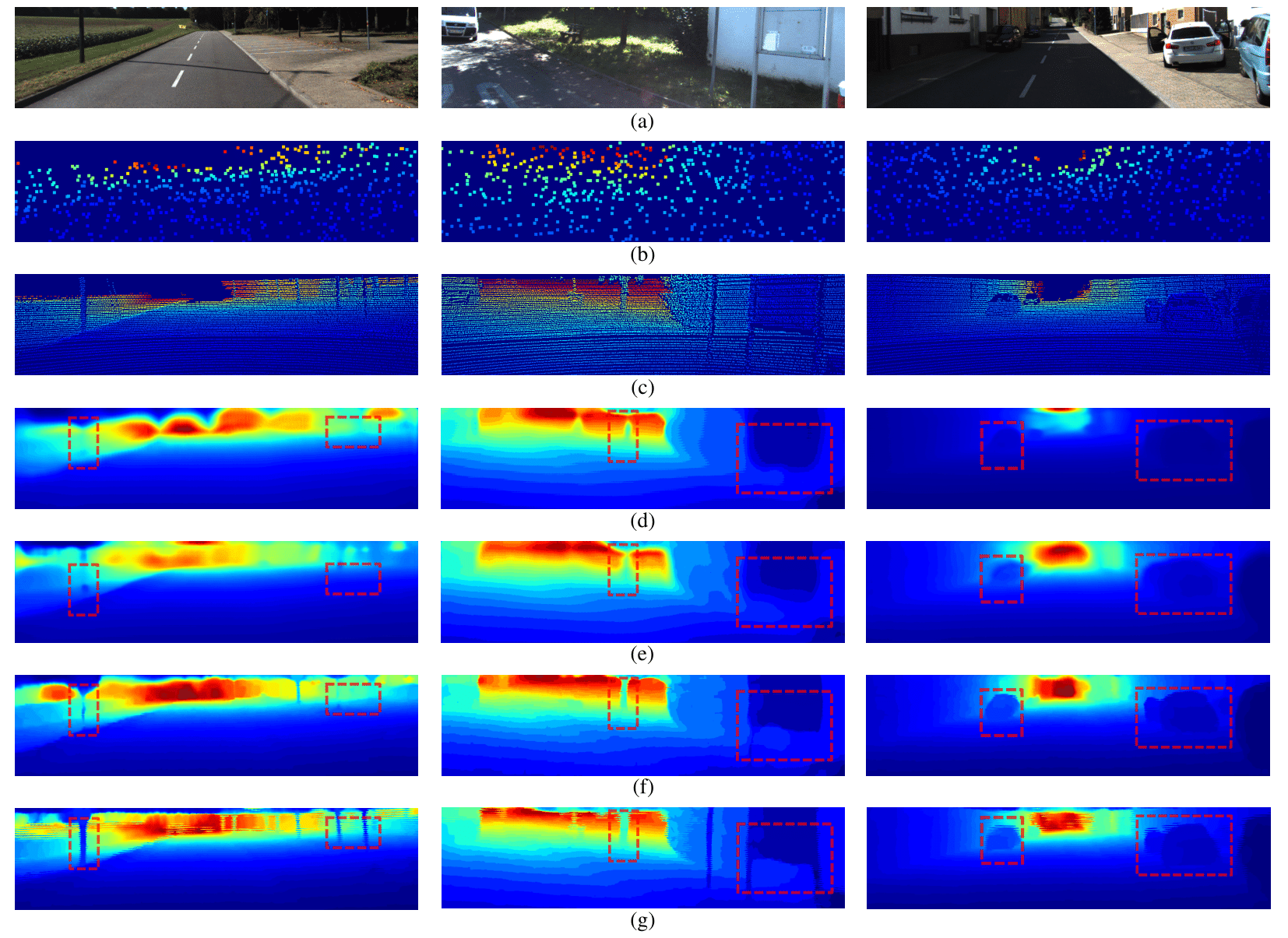}
\caption{Qualitative comparisons on KITTI dataset. (a) Input image; (b) Sparse depth samples(500); (c) Ground Truth; (d) Ma \etal \cite{Ma2017SparseToDense}; (e) Ma\cite{Ma2017SparseToDense}+SPN\cite{liu2016learning}; (f) MC+CSPN(Ours);(g) MC+CSPN+ACSPF(Ours). Most significantly improved regions are highlighted with dash boxes (best view in color).}
\label{fig:kitti}
\vspace{-1.0\baselineskip}
\end{figure*}

\begin{table*}[t]
\centering
\caption{Ablation studies for 3D module on the scene flow dataset~\cite{mayer2016large}.}
\label{tbl:3d_ab_study}
\fontsize{8.5}{8.5}\selectfont
\bgroup
\def\arraystretch{1.4}
\setlength{\tabcolsep}{6pt} 
\begin{tabular}{lcccccc} 
    \toprule
    \multirow{2}{*}{Method}       & \multicolumn{1}{l}{\multirow{2}{*}{}} & \multicolumn{2}{c}{~~Lower the Better~~} &                      & \multicolumn{2}{c}{~~CSPN Parameters~~}                              \\ 
    \cline{3-4}\cline{6-7}
                                  & \multicolumn{1}{l}{}                  & ~~EPE~~            & ~~RMSE~~                & \multicolumn{1}{l}{} & ~ ~ Propagation Times~ ~ ~    & \multicolumn{1}{l}{~~Kernel Size~~}  \\ 
    \hline
    ~~ PSMNet~\cite{chang2018pyramid}                       &                                       & 1.183          & 5.680               &                      & 0                             & 0                                \\
    w CSPN                         &                                       & 0.992          & 5.142               &                      & 24                            & 3                                \\
    w 3DCSPN\_ds                   &                                       & 0.971          & 5.129               &                      & 24                            & 3                                \\
    w 3DCSPN\_ss                   &                                       & 1.007          & 4.731               &                      & 1                             & 3                                \\
    w 3DCSPN\_ds\_ss               &                                       & \textbf{0.951} & \textbf{4.561}      &                      & 24(ds)+1(ss)                  & 3                                \\ 
    \hline
    w 3DCSPN\_ds\_ss + ACSPF &                                       & \textbf{0.777} & \textbf{4.352}      &                      & 1(ACSPP)+1(CFF)+24(ds)+1(ss) & 3                                \\
    \bottomrule
\end{tabular}
\vspace{-1.0\baselineskip}
\egroup
\end{table*}

\subsubsection{Comparisons}
\label{subsec:compare}

\peng{\noindent\textbf{Baseline algorithms.}} 
We compare our method against one of the SOTA algorithms for depth completion, \ie Sparse-to-Dense~\cite{Ma2017SparseToDense}, and against the most related SOTA dense prediction refinement strategies, \ie~SPN~\cite{liu2016learning}, to validate our algorithm.

Specifically, we adopt exactly the same backbone as~\cite{Ma2017SparseToDense}, and modify only the architecture with our mirror connection (MC) (\secref{subsec:unet}), and add additional module of 2D CSPN as illustrated in ~\figref{fig:arch_single}. 
To compare against SPN~\cite{liu2016learning}, we use the same feature map for affinity prediction, while only switch the linear propagation strategy. For reproducing the baselines, we use released code online from the authors.

In the following, as shown in ~\tabref{tbl:sota} $\&$~\tabref{tbl:sota_kitti}, we compare  various methods in two different settings. (1) Direct refine the depth map after the network output without including sparse depth samples in propagation, \ie not preserve sparse samples (SD) in final outputs. (2) Refine the depth using both the network output and sparse depth samples, \ie preserve SD. In both cases, we show our proposed approaches largely outperform the baselines, yielding SOTA results for depth completion with very sparse depth samples.

\noindent\textbf{NYU v2.} \tabref{tbl:sota} shows the comparison results. Our baseline results are the depth output from the network of Ma \etal~\cite{Ma2017SparseToDense}, which takes as input a RGB image and a sparse depth map.
At the upper part of \tabref{tbl:sota} we show that the results for depth refinement using network outputs only.
At row ``Bilateral'', we refine the network output from~\cite{Ma2017SparseToDense} using bilateral filtering~\cite{barron2016fast} as a post-processing module with their spatial-color affinity kernel tuned on our validation set. Although the output depths snap to image edges, the absolute depth accuracy is dropped since the filtering over-smoothed original depths. At row ``DenseCRF``, we show the results filtered with DenseCRF~\cite{philipp2012dense}. Specifially, we first evenly discretized the depth value to $256$ labels, and then did post-processing with both spatial and color RBF kernels with tuned parameters. The estimated depths are significantly improved thanks to the designed affinity.
At  row ``SPN'', we show the results filtered with SPN~\cite{liu2017learning}, using the author provided affinity network. Due to joint training, the depth is improved with the learned affinity, yielding both better depth details and absolute accuracy. Switching SPN to CSPN (row ``CSPN'') yields significantly better results, \eg~accuracy of $\delta \leq $ 1.02 increases from 67.5$\%$ to 76.2$\%$, demonstrating its effectiveness.
Finally, at the row ``Mirror connection (MC)'', we show the results of just modifying the network with mirror connections as stated in~\secref{subsec:unet}.  The results turn out to be even better than that from SPN and CSPN, demonstrating that by simply adding feature from the initial layers, depth can be better learned.

At the lower part of \tabref{tbl:sota}, we show the results using both network outputs and sparse depth samples at later stage, so that all the results preserves the sparse depth value provided. We randomly select 500 depth samples per image from the ground truth depth map. 
At row ``Replacement'', as illustrated in \figref{fig:gradient}(c), we first directly replace the depth values at sparse depth sample locations, yielding reasonable improvements in depth accuracy. However, the visual results are not smooth, which is not satisfactory. In addition, we consider a baseline method using as-rigid-as-possible (ARAP)~\cite{igarashi2005rigid} warping. Basically the input depth map is warped with the sparse depth samples as control points. At row ``ARAP'', we show its results, which just marginally improves the estimation over the baseline network. 
For SPN, we also apply the similar replacement operation in \equref{eqn:cspn_sp} for propagation, and the results are shown at row ``SPN'', which outperforms both the results form ARAP and SPN without propagation of SD due to joint training helps fix the error of warping.  At row ``MC + SPN'', we use our architecture with mirror connection for learning affinity with SPN, which outperforms ``SPN'', while we did not see any improvements compared with that only using MC. 
Nevertheless, by replacing SPN with our CSPN, as shown in row ``MC + CSPN'', the results can be further improved by a large margin and performs best in all cases. We think this is mostly because CSPN updates more efficiently than SPN during the training. 

Finally, we validate the effect of our convolutional spatial pyramid fusion (CSPF) (\secref{subsec:cspn_var}). We first add ASPP module in middle part of the network (``MC+CSPN+ASPP'') to serve as the baseline, which slightly improves the results. Then we add convolutional feature fusion (CFF) (``MC+CSPN+ASPP+CFF''), and add repalce ASPP with ACSPP (``MC+CSPN+ACSPP''), showing both of the components proposed in CSPF are effective for improving the performance. Jointly using ACSPP and CFF (``MC+CSPN+ACSPF'') yields the best performance. This is majorly because larger context and better spatial details are obtained.

We provide more complete ablation validation of CSPF in~\secref{subsec:stereo_exp} due to PSMNet also adopts SPP in their architecture.
Some visualizations are shown in \figref{fig:nyudepth}. We found the results from CSPN and CSPF do capture better structure from images (highlighted with dashed bounding boxes) than that from other state-of-the-art strategies.

\noindent\textbf{KITTI.} \tabref{tbl:sota_kitti} shows the depth refinement with both color and sparse depth samples. Ours final model ``MC + CSPN + ACSPF'' largely outperforms other SOTA strategies, which shows the generalization of the proposed approach. For instance, with a very strict metric $\delta < 1.02$, ours improves the baseline~\cite{Ma2017SparseToDense} from $30\%$ to $72\%$. More importantly, CSPN is running very efficiently, thus can be applied to real applications.
Some visualization results are shown at the bottom in \figref{fig:kitti}. Compared to the network outputs from~\cite{Ma2017SparseToDense} and SPN refinement, CSPN sees much more details and thin structures such as poles near the road (first image (f)), and trunk on the grass (second image (f)). For the third image, we highlight a car under shadow in the left, whose depth is difficult to learn. We can see SPN fails to refine such a case in (e) due to globally varying lighting variations, while CSPN learns local contrast and successfully recover the silhouette of the car. Finally, by adding ``ACSPF'' in the framework, the spatial reduction of last block in ResNet encoder is removed (as shown in \figref{fig:arch_single}). Therefore, the results are further improved, and better local details are recovered.
Finally, we also submit our results to the KITTI depth completion challenge~\footnote{\url{http://www.cvlibs.net/datasets/kitti/eval_depth.php?benchmark=depth_completion}} and show that our results is better than previous SOTA method~\cite{uhrig2017sparsity} at the time of submission.

\begin{table*}[!htpb]
\centering
\caption{Ablation studies for various spatial pyramid module on scene flow dataset~\cite{mayer2016large}.}
\label{tbl:cpm_study}
\fontsize{8.5}{8.5}\selectfont
\bgroup
\def\arraystretch{1.4}
\setlength{\tabcolsep}{6pt} 
\begin{tabular}{lcccccccc} 
\toprule
\multicolumn{1}{l}{\multirow{2}{*}{Method}} & \multirow{2}{*}{}    & \multicolumn{2}{c}{~~Lower the Better~~} &  & \multicolumn{4}{c}{~~Addtional Setting~~}                                  \\ 
\cline{3-4}\cline{6-9}
\multicolumn{1}{c}{}                        &                      & ~~EPE~~  & ~~RMSE~~                                                                            &  & ~~3DCSPN\_ds~~ & \multicolumn{1}{l}{~~2DCSPN~~} & ~~3DCSPN\_Fusion~~ & ~~Dilation~~  \\ 
\hline
SPP (PSMNet~\cite{chang2018pyramid})                                 &                      & 1.183 & 5.680                                                                           &  &             & \multicolumn{1}{l}{}       &                 &           \\
SPP                                         & \multicolumn{1}{l}{} & 0.971 & 5.129                                                                           &  & \checkmark           &                            &                 &           \\
CSPP                                        &                      & 0.954 & 5.184                                                                           &  & \checkmark           & \checkmark                          &                 &           \\
ASPP                                        &                      & 0.970 & 5.165                                                                           &  & \checkmark           &                            &                 & \checkmark         \\
ACSPP                                       &                      & 0.902 & 4.954                                                                           &  & \checkmark           & \checkmark                          &                 & \checkmark         \\
SPP+CFF                               &                      & 0.905 & 5.036                                                                           &  & \checkmark           &                            & \checkmark               &           \\
ACSPF                                  &                      & \textbf{0.827} & \textbf{4.555}                                                                           &  & \checkmark           & \checkmark                          & \checkmark               & \checkmark         \\
\bottomrule
\end{tabular}
\egroup
\end{table*}

\subsection{Stereo depth estimation}
\label{subsec:stereo_exp}

\noindent\textbf{Implementation details.}
The base network we adopted is from the PSMNet~\cite{chang2018pyramid}, and we follow the same training strategies. Specifically, for learning CSPN, we adopt Adam ~\cite{kingma2014adam} optimizer with $\beta_1$ = 0.9, $\beta_2$ = 0.999, and batch size is set to 16 for training on eight Nvidia P40 GPUs(each of 2). 
We performed color normalization on the entire dataset of Scene Flow for data preprocessing. During training, we crop the input image to $512 \times 256$. We first train our network from scratch on Scene Flow dataset for 10 epochs, and the learning rate during this periods is set to $0.001$. When train on KITTI, 
we finetune the model obtained from Scene Flow for another 600 epochs. The learning rate starts from $0.001$ and decrease 10\% each 200 epochs. Acquired by accumulating Velodyne HDL-64 Laser scanner, KITTI stereo ground truth is relatively sparse, and we only compute the loss where LiDAR ground truth is available. 

\subsubsection{Datasets and Metrics}
\label{subsec:data_metric}

we evaluate our method on following datasets: Scene Flow~\cite{mayer2016large}, KITTI Stereo 2012~\cite{geiger2012we}, KITTI Stereo 2015~\cite{Menze2018JPRS}.

\noindent\textbf{Scene Flow.} A large scale dataset contains 35454 training and 4370 test stereo pairs in 960x540 pixel resolution, rendered from various synthetic sequences. Pixels besides our max disparity are excluded in loss function.

\noindent\textbf{KITTI Stereo 2012.} A real-world dataset with street views from a driving car, consists of 194 training and 195 test stereo pairs in 1240x376 resolution. Ground truth has been aquired by accumulating 3D point clouds from a 360 degree Velodyne HDL-64 Laserscanner. We divided the whole training data into 160 training and 34 validate stereo pairs, we adopted color image as network input in this work.

\noindent\textbf{KITTI Stereo 2015.} Compared to KITTI 2012, KITTI 2015 consists of 200 training and 200 test stereo pairs in 1240x376 resolution. Also, it comprises dynamic scenes for which the ground truth has been established in a semi-automatic process. We further divided the whole training data into 160 training and 40 validate stereo pairs.

\noindent\textbf{Metrics.} Since different datasets have various metrics for comparison, we list the corresponding evaluation metric as follows,

Scene Flow: the end-point error (EPE) is used. Formally, the difference could be written as $EPE(d^*, \hat{d}) = \|d^* - \hat{d}\|_2$.

KITTI 2012 \textcolor{black}{and} 2015: the percentages of erroneous pixels. Specifically, a pixel is considered to be an erroneous pixel when its disparity error is larger than $t$ pixels. Then, the percentages of erroneous pixels in non-occluded (Out-Noc) and all (Out-All) areas are calculated. Specifically, for benchmark 2012, $t \in \{2, 3, 4, 5\}$. While for benchmark 2015, a pixel is considered to be wrong when the disparity error is larger than $3$ pixels or relatively $5\%$, whichever is less stringent. In addition, results on both left image (D1-All) and right image (D2-All) are evaluated. We refer the reader to their original page~\footnote{\url{http://www.cvlibs.net/datasets/kitti/eval_stereo_flow.php?benchmark=stereo}} for more detailed information about other evaluated numbers. Here we only list the major metric to rank different algorithms.

\begin{table*}[!htpb]
	\centering
	\vspace{-1\baselineskip}
	\caption{Results on Scene Flow dataset and KITTI Benchmarks.}
	\label{tbl:benchmark}
	\fontsize{7.5}{8}\selectfont
	\bgroup
	\def\arraystretch{1.3}
	\setlength{\tabcolsep}{4.5pt} 
    \begin{tabular}{lc|cccccccc|cc} 
    \toprule
    \multirow{3}{*}{Method}     & \multicolumn{1}{c}{Scene Flow} & \multicolumn{8}{c}{KITTI 2012}                                                                                                         & \multicolumn{2}{c}{KITTI 2015}   \\ 
    \cline{2-12}
                                & \multirow{2}{*}{EPE} & \multicolumn{2}{c}{2px}         & \multicolumn{2}{c}{3px}         & \multicolumn{2}{c}{4px}         & \multicolumn{2}{c|}{5px}        & All            & Non-occluded    \\
                                &                      & Out-Noc        & Out-All        & Out-Noc        & Out-All        & Out-Noc        & Out-All        & Out-Noc        & Out-All        & D1-all         & D1-all          \\ 
    \hline
    MC-CNN~\cite{zbontar2016stereo}     & 3.79                 & 3.90           & 5.45           & 2.43           & 3.63           & 1.90           & 2.85           & 1.64           & 2.39           & 3.88           & 3.33            \\
    SGM-Net~\cite{seki2017sgm}    & 4.50                 & 3.60           & 5.15           & 2.29           & 3.50           & 1.83           & 2.80           & 1.60           & 2.36           & 3.66           & 3.09            \\
    Displets v2~\cite{guney2015displets} & 1.84                 & 3.43           & 4.46           & 2.37           & 3.09           & 1.97           & 2.52           & 1.72           & 2.17           & 3.43           & 3.09            \\
    GC-Net~\cite{kendall2017end}     & 2.51                 & 2.71           & 3.46           & 1.77           & 2.30           & 1.36           & 1.77           & 1.12           & 1.46           & 2.67           & 2.45            \\
    iResNet-i2~\cite{liang2017learning} & 1.40                 & 2.69           & 3.34           & 1.71           & 2.16           & 1.30           & 1.63           & 1.06           & 1.32           & 2.44           & 2.19            \\
    PSM-Net~\cite{chang2018pyramid}    & 1.09                 & 2.44           & 3.01           & 1.49           & 1.89           & 1.12           & 1.42           & 0.90           & 1.15           & 2.32           & 2.14            \\
    EdgeStereo~\cite{song2018edgestereo} & 1.12                 & 2.79           & 3.43           & 1.73           & 2.18           & 1.30           & 1.64           & 1.04           & 1.32           & 2.16           & 2.00            \\ 
    \hline
    w 3DCSPN\_ds\_ss (Ours)      & 0.95      & 1.95  & 2.47  & 1.25  & 1.61  & 0.96  & 1.23  & 0.79  & 1.00  & 1.93  & 1.77   \\
    w 3DCSPN\_ds\_ss + CSPF (Ours)    & \textbf{0.78}        & \textbf{1.79}  & \textbf{2.27}  & \textbf{1.19}  & \textbf{1.53}  & \textbf{0.93}  & \textbf{1.19}  & \textbf{0.77}  & \textbf{0.98}  & \textbf{1.74}  & \textbf{1.61}   \\
    \bottomrule
    \end{tabular}
	\egroup
	\vspace{-1\baselineskip}
	\label{tbl:benchmark}
\end{table*}

\subsubsection{Ablation Study}
We do various ablation studies based on the Scene Flow dataset to validate each component of our networks as shown in \figref{fig:arch} and \figref{fig:CPM}. 
We first train a baseline results with the code provided online by the author of PSMNet~\footnote{\url{https://github.com/JiaRenChang/PSMNet}}. 

\textbf{Study the 3D module.} We first evaluate the components proposed in our 3D module (\figref{fig:arch}) in \tabref{tbl:3d_ab_study}. 
In order to show that propagation in the new dimension benefits the results, we first adopt 2D CSPN for depth refinement as proposed for single image depth refinement over the three 2D disparity maps using the affinity predicted from the same feature, \ie~``CSPN''. As expected, it reduces the EPE error from 1.119 to 0.992. Then, we switch the 2D CSPN to 3D CSPN as proposed in \secref{subsec:stereo},  \ie~``3DCSP\_{ds}'', the results are further improved to 0.971. Here, the footnote ``ds'' is short for disparity space, indicating the 3DCSPN is performed over the disparity outputs with shape $d \times h \times w \times 1$. ``3DCSPN\_{ss}'' shows the results by using 3D CSPN over the space for multi-stage outputs fusion, which also helps the performance from our baseline.
Jointly using the two 3D CSPNs, \ie ``3DCSPN\_{ds}\_{ss}'', yields the best result, outperforming our baseline method by a large margin. 
At last row, ``3DCSPN\_{ds}\_{ss} + ACSPF'' shows the results of combining our 3D module with our enhanced ASPP module together, which additionally reduce the error around $30\%$ \wrt to the baseline.

\textbf{Study the CSPF module.}
Here, we evaluate different components for enhancing the SPP module that is also adopted in PSMNet, as shown in Fig. \ref{fig:CPM}. For all the variations, we adopt ``3DCSPN\_{ds}'' as our 3D module for ablation study. As introduced in \secref{subsec:cspn_var}, ``CSPP'' means we use 2D CSPN over the spatial pooling grid, which reduces the EPE error from 0.971 to 0.954. We then study another spatial pooling strategy with dilated convolution, \ie ``ASPP'', which produces similar performance as SPP. Surprisingly, as shown in row ``ACSPP'', jointly using our 2D CSPN with ASPP produces error much smaller (0.902) than that with SPP (0.954). At row ``CFF'', we use out proposed fusion strategy to combine the pooled features from the spatial pyramid, which also significantly improves over the SPP baseline, reducing EPE error from 0.954 to 0.905. 
Finally, combining ACSPP and CFF, \ie ``ACSPF'', yields the best performance, which is selected as our final SPP module.
\figref{fig:res}(c) shows a few examples of the output from the Scene Flow dataset, and we can see the predicted results are very close to ground truth, which are exceptionally good in handling detailed object structures.

\subsubsection{Comparisons}

\peng{\noindent\textbf{Baseline algorithms.}}
To validate the algorithm, in addition to comparing over the Scene Flow test set, we also submitted our results to KITTI 2012 and 2015 test evaluation server to compare against other SOTA methods proposed in recent years, including PSM-Net~\cite{chang2018pyramid}, iResNet-i2~\cite{liang2018learning}, GC-Net~\cite{kendall2017end}, EdgeStereo~\cite{song2018edgestereo}, SGMNet~\cite{seki2017sgm}, Displets v2~\cite{guney2015displets} and MC-CNN~\cite{zbontar2016stereo}. 
\textcolor{black}{For KITTI benchmarks, we choose the same maximum disparity value used in PSM-Net~\cite{chang2018pyramid}, which is 192.} 

\begin{figure*}[!htpb]
	\centering
	\includegraphics[width=1.0\textwidth]{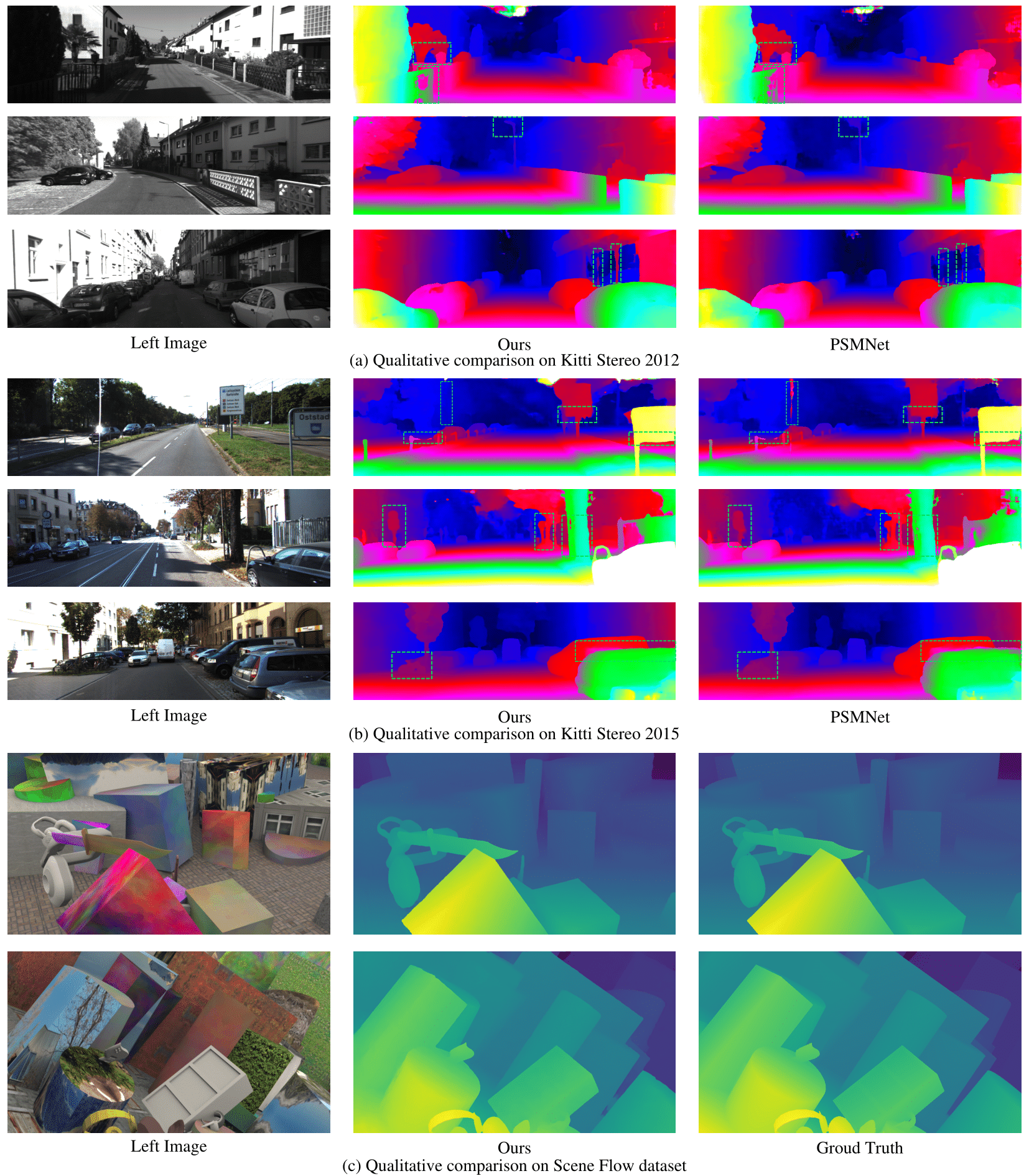}
	\caption{ Qualitative results. By learning affinity matrix in our model and propagate it to leverage context better, we can handle more challenging case. Significantly improved regions are highlight with green dash boxes(best view in color).}
	\label{fig:res}
\end{figure*}

As summarized in Table \ref{tbl:benchmark}, our method outperforms all others methods by a notable margin (above relatively $10\%$), and performs the best over all the major metrics both in KITTI 2012~\footnote{\url{http://www.cvlibs.net/datasets/kitti/eval_stereo_flow.php?benchmark=stereo}} and 2015~\footnote{\url{http://www.cvlibs.net/datasets/kitti/eval_scene_flow.php?benchmark=stereo&eval_gt=all&eval_area=all}}. By checking detailed numbers in KITTI 2015, we are better at improving the static background than the foreground, which is reasonable because background has much larger amount of training pixels for learning the propagation affinity.
\figref{fig:res} (a) and (b) show several examples by comparing our algorithm to the baseline method PSMNet over KITTI 2012 and 2015 respectively, and we mark out the improved regions with dashed bounding boxes. As can be seen, CSPN not only better recovers the over all scene structure, but also superior in recovering detailed scene structures. More results are available in the KITTI leaderboard pages.

\section{Conclusion}
In this paper, we propose an effective module, namely convolutional spatial propagation network (CSPN), for two depth estimation tasks, \ie~depth completion and stereo depth estimation. CSPN can be jointly learned with any type of depth estimation neural networks, and could be regarded as a linear diffusion process with guarantee of model stability. Comparing with previous spatial propagation network~\cite{liu2017learning}, CSPN is more efficient (2-5$\times$ faster in practice especially when the image is large), and more accurate (over $30\%$ improvement) in terms of depth completion. 

 In addition, we extend CSPN to 3D, namely~3D CSPN, and combine it with spatial pyramid pooling (SPP), namely CSPF. These modules are shown to be effective in the two tasks we focused.  
 Last, we further tune CSPN to better adapt each task. For depth completion, we embed sparse depth samples into its propagation process, and for stereo matching, we allow 3D CSPN to diffuse along the dimension of both the disparity spaces and the scale space. 
For both tasks, our designed approaches with CSPN provides superior improvement over other SOTA methods~\cite{Ma2017SparseToDense, chang2018pyramid}. Since many components of our framework are general, in the future, we plan to apply them to other tasks such as image segmentation and image enhancement.

\IEEEdisplaynontitleabstractindextext

%
\IEEEpeerreviewmaketitle

\ifCLASSOPTIONcompsoc
  \section*{Acknowledgments}
\else
  \section*{Acknowledgment}
\fi
This work is supported by Baidu Inc.

\ifCLASSOPTIONcaptionsoff
  \newpage
\fi



\bibliographystyle{IEEEtran}
\bibliography{IEEEabrv,egbib}
\end{document}